%% file: main.tex
\documentclass[10pt,twocolumn,letterpaper]{article}

\usepackage[pagenumbers]{cvpr}

\input{preamble}
\definecolor{cvprblue}{rgb}{0.21,0.49,0.74}
\usepackage[pagebackref,breaklinks,colorlinks,allcolors=cvprblue]{hyperref}

\usepackage{multirow}
\definecolor{myred}{rgb}{0.996,0.578,0.574}
\definecolor{myyellow}{rgb}{0.988,0.961,0.898}
\definecolor{mylightred}{rgb}{0.992,0.887,0.883}
\usepackage{array}
\usepackage{booktabs}
\usepackage{caption}
\usepackage[flushleft]{threeparttable}
\usepackage{colortbl}
\usepackage[table,xcdraw]{xcolor}
\usepackage{tcolorbox}
\usepackage{caption}
\usepackage{array}
\usepackage{makecell}
\definecolor{cherryred}{RGB}{220,20,60}
\definecolor{our_main_color}{RGB}{217, 242, 208}

\def\paperID{******}
\def\confName{CVPR}
\def\confYear{2026}

\title{OralGPT-Plus: Learning to Use Visual Tools via Reinforcement Learning for Panoramic X-ray Analysis}

\author{%
    Yuxuan Fan\textsuperscript{1,3}\thanks{Equal contribution.}\quad
    Jing Hao\textsuperscript{2}$^\ast$$^\spadesuit$\quad
    Hong Chen\textsuperscript{1}\quad
    Jiahao Bao\textsuperscript{4}\quad
    Yihua Shao\textsuperscript{5}\quad \\
    Yuci Liang\textsuperscript{6}\quad
    Kuo Feng Hung\textsuperscript{2}\quad
    Hao Tang\textsuperscript{3}\thanks{Corresponding authors: haotang@pku.edu.cn}\quad \\
    \textsuperscript{1}The Hong Kong University of Science and Technology (GZ)\quad \\
    \textsuperscript{2}Faculty of Dentistry, The University of Hong Kong\quad \\
    \textsuperscript{3}School of Computer Science, Peking University\quad \\
    \textsuperscript{4}Shanghai Jiao Tong University\quad \\
    \textsuperscript{5}Institute of Automation, Chinese Academy of Sciences\quad \\
    \textsuperscript{6}College of Computer Science and Software Engineering, Shenzhen University\quad \\
}

\begin{document}
\maketitle

\begingroup
\renewcommand\thefootnote{}
\footnotetext{\hspace*{-0.4em}$^\spadesuit$ Project Leader.}
\endgroup

\input{0_abstract}    
\input{1_intro}
\input{2_preliminary}
\input{3_methods}

\input{4_benchmark}

\input{5_experiments}
\input{6_conclusion}
{
    \small
    \bibliographystyle{ieeenat_fullname}
    \bibliography{main}
}

\clearpage
\input{X_suppl}

\end{document}

%% file: 0_abstract.tex
\begin{abstract}
Panoramic dental radiographs require fine-grained spatial reasoning, bilateral symmetry understanding, and multi-step diagnostic verification, yet existing vision–language models operate under a static single-pass paradigm that limits their clinical reliability. In this paper, we introduce OralGPT-Plus, an agentic vision–language model designed to perform iterative and symmetry-aware diagnostic reasoning for panoramic dental radiograph analysis. To support this paradigm, we construct DentalProbe, a five-thousand–image dataset with expert-curated diagnostic trajectories that provide structured supervision for localized inspection and contralateral comparison. We further develop a Reinspection-driven reinforcement learning framework that encourages clinically meaningful re-examination and stabilizes long-horizon reasoning with rubric-based reward and conditioned diagnostic-driven reward. In parallel, we present MMOral-X, the first benchmark for holistic panoramic diagnosis, containing 300 open-ended questions and region-level annotations across multiple difficulty levels. OralGPT-Plus demonstrates consistent and reliable improvements over strong baselines on MMOral-X and established panoramic benchmarks, indicating the effectiveness of interactive and symmetry-informed reasoning. Our work highlights the value of agentic modeling for dental imaging and provides a foundation for future research in clinically aligned panoramic radiograph analysis. Codes will be published in \url{https://github.com/isbrycee/OralGPT}.
\end{abstract}

%% file: 1_intro.tex
\begin{figure}[t]
  \centering
\includegraphics[width=1\linewidth]{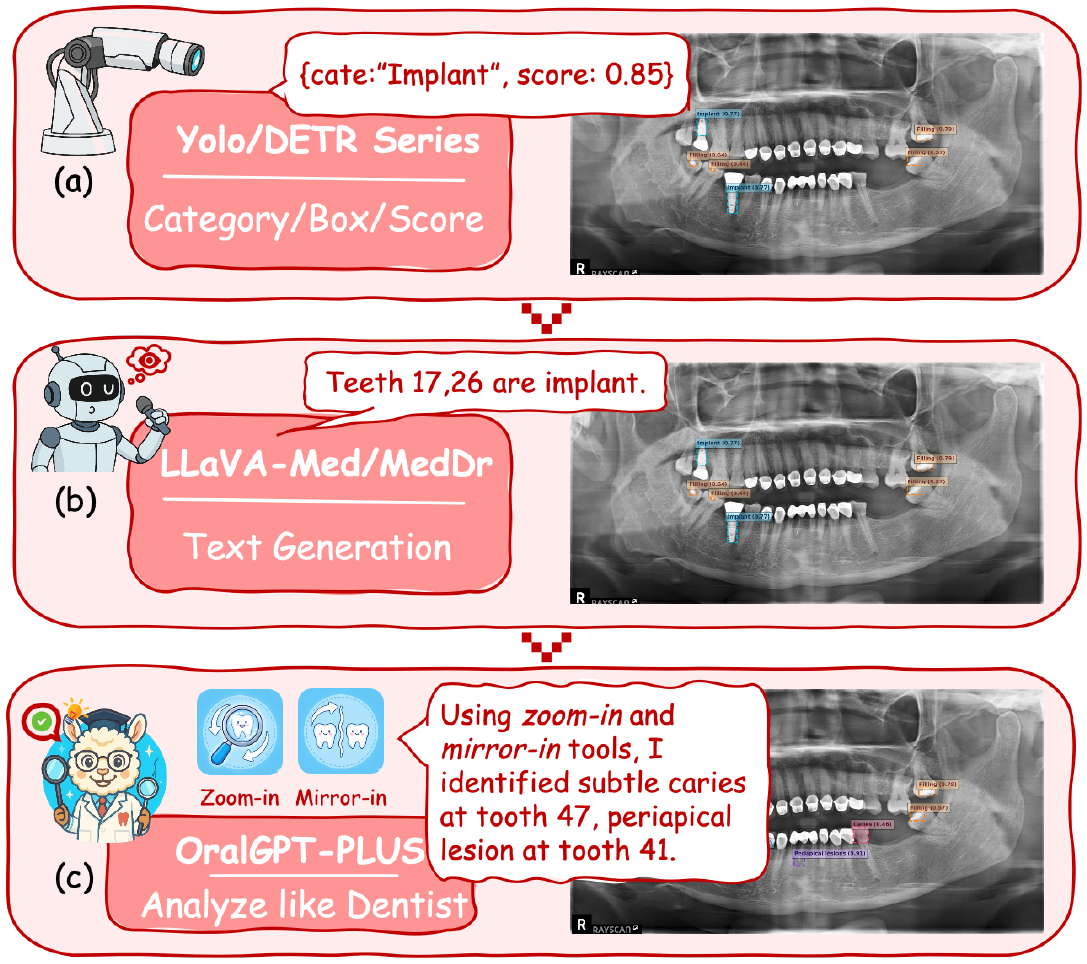}
  \caption{\textbf{Paradigm shift in panoramic dental radiograph analysis.} Traditional detectors provide only category-level boxes, while VLMs generate coarse descriptions without structured reasoning. In contrast, OralGPT-Plus performs dentist-like diagnostic reasoning by invoking zoom-in and mirror-in tools to identify subtle findings and produce clinically coherent interpretations.}
  \label{fig:3_paradigm}
  \vspace{-0.4cm}
\end{figure}

\section{Introduction}
Panoramic radiograph is universally recognized as a fundamental tool for oral disease diagnosis and treatment planning, as it provides a comprehensive view of the teeth, alveolar bone, and surrounding anatomical structures within a single image~\cite{hao2024semi,hao2024semit,hao2025characteristics,wang2026miccai}. An exhaustive analysis of panoramic radiographs is indispensable for both screening and therapeutic decision-making~\cite{liu2025performance1}. However, interpreting panoramic radiographs is extremely time-consuming and often relies heavily on the experience of radiologists~\cite{hao2024semi,hao2024semit,hao2025characteristics}. The computer vision techniques have promoted the advancement of intelligent dentistry~\cite{hao2024tmamba,hao2025towards}. Figures~\ref{fig:3_paradigm}(a) and Figures~\ref{fig:3_paradigm}(b) show two widely used anomaly detection paradigms. Traditional object detectors~\cite{hao2025gem,hao2024language} focus on localizing suspected abnormalities and return only categories, bounding boxes, and confidence scores, limiting their ability to support clinically meaningful explanations. Vision-language models (VLMs)~\cite{Li2023LLaVAMedTA,hao2025towards,hao2025oralgpt, cai2025dentalgpt,he2024meddr,sun2024descriptive,hao2024fullanno} have improved semantic expressiveness; yet, the single forward prediction paradigm restricts their ability to revisit ambiguous regions and capture subtle disease patterns.
Unlike other dental imaging modalities, panoramic radiographs have a higher image resolution, making dentists frequently magnify areas suspected of containing abnormalities for closer examination. Additionally, they often refer to analogous teeth in other quadrants due to dental symmetry. By comparing corresponding teeth in symmetrical quadrants, they can perceive and extract subtle features, thus identifying minute disease patterns such as carious lesions, apical periodontitis, and others~\cite{du2024prompting}.
Inspired by this clinical and practical workflow, we propose OralGPT-Plus, an agentic model that performs diagnostic reasoning through an iterative cycle of thought, action, and observation. The model is equipped with a focused inspection operator that produces localized high-resolution views of suspicious areas, as well as a symmetry-aware operator that retrieves the corresponding anatomical region on the opposite side. These abilities enable the model to refine ambiguous regions, perform bilateral comparisons when needed, and progressively form coherent diagnostic conclusions. This iterative reasoning mirrors core elements of clinical workflows rather than producing a single, static summary.

Developing such a model requires supervision that reflects real diagnostic behavior. For this purpose, we curate DentalProbe, a dataset that includes more than eight thousand rounds of expert-guided diagnostic trajectories. The trajectories capture the sequence of global inspection, proposal identification, localized examination, and symmetry-based verification that characterizes panoramic radiograph interpretation. Figure~\ref{fig:data_curation} illustrates the curation pipeline, which integrates multiple public datasets with expert annotations and multi-agent trajectory refinement. This corpus enables instruction tuning that teaches the model to follow clinically consistent reasoning patterns and to use the diagnostic tools appropriately.

Although instruction tuning provides a foundation for structured behavior, effective multi-turn analysis benefits from further optimization. We therefore introduce a reinforcement learning framework that encourages the model to re-examine regions when necessary and limit unproductive exploratory actions. The reward design offers continuous feedback that reflects diagnostic completeness and accuracy, which alleviates the sparsity that commonly arises in multi-lesion scenarios. It also regulates the initiation of additional inspections so that the model focuses on clinically meaningful reasoning. These mechanisms contribute to stable long-horizon optimization and support balanced and reliable decision-making.

To evaluate the diagnostic performance in a comprehensive manner, we construct MMOral-X, a benchmark that contains 300 open-ended question-answer pairs covering clinically relevant diseases and findings. The benchmark includes three levels of difficulty that reflect the number, variety, and visual ambiguity of potential abnormalities. It emphasizes holistic analysis from a single panoramic radiograph and enables systematic assessment of multi-turn reasoning, fine-grained lesion interpretation, and symmetry-aware analysis.

Across all components of the framework, OralGPT-Plus pursues interpretability, stability, and alignment with clinical expectations. The experimental results show consistent improvements over strong baselines on MMOral-X, as well as on one established public benchmark. The gains are steady rather than abrupt and reflect the value of interactive and symmetry-aware reasoning strategies in panoramic dental radiograph analysis.

\textbf{Our contributions are summarized as follows:}
\begin{itemize}
    \item We introduce OralGPT-Plus, an agentic and symmetry-aware model designed to perform iterative diagnostic reasoning on panoramic dental radiographs.
    \item We curate DentalProbe, a five-thousand image dataset with expert diagnostic trajectories that support clinically aligned instruction tuning and structured tool usage.
    \item We present MMOral-X, a benchmark for holistic panoramic radiograph interpretation that evaluates holistic understanding across varying levels of complexity.
    \item Extensive evaluations show that OralGPT-Plus achieves strong and consistent performance, indicating that interactive and symmetry-informed reasoning substantially improves diagnostic quality.
\end{itemize}

%% file: 2_preliminary.tex
\section{Preliminary}
\label{sec:preliminary}

\begin{figure*}[!ht]
  \centering
  \includegraphics[width=\textwidth]{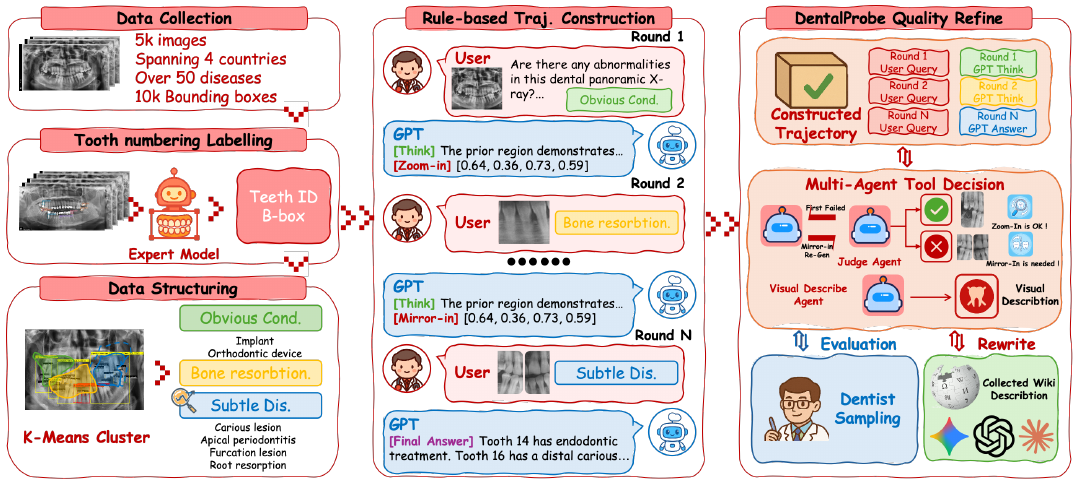}
  \caption{\textbf{Curation process of DentalProbe dataset with expert trajectory for dentist-like instruction tuning.}}
  \vspace{-0.4cm}
  \label{fig:data_curation}
\end{figure*}

\begin{figure}[!ht]
  \centering
  \includegraphics[width=0.8\linewidth]{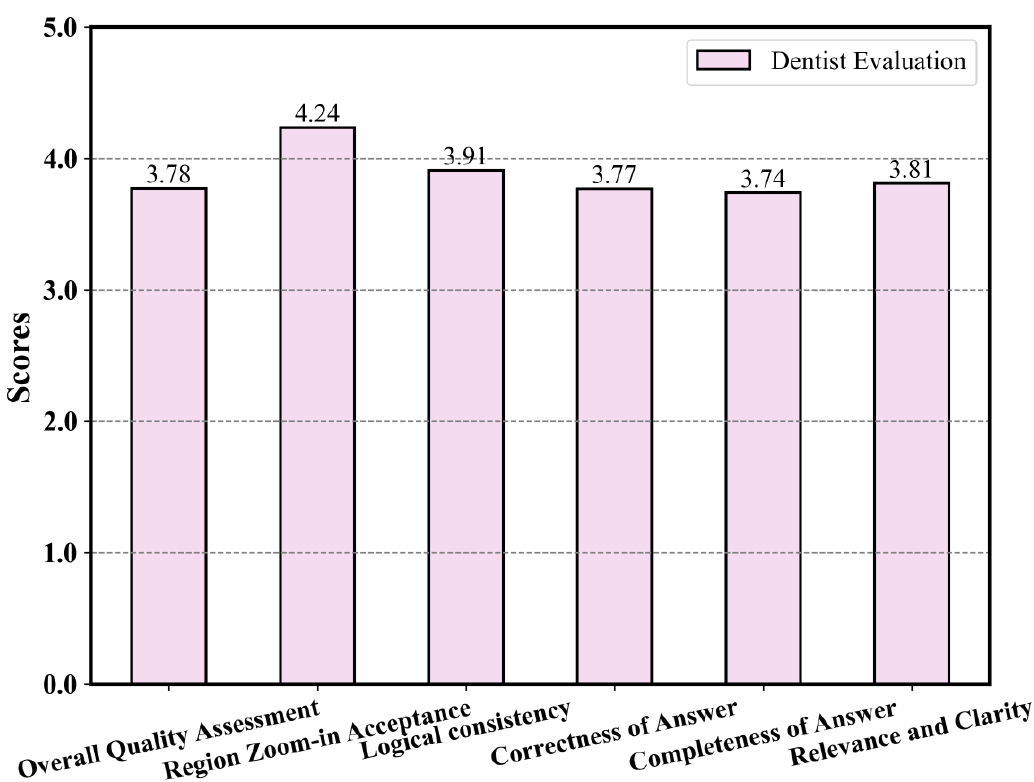}
  \caption{\textbf{Average score under dentist sampling evaluation on curated DentalProbe dataset.}}
\vspace{-0.4cm}
  \label{fig:dentist_evaluation}
\end{figure}

Fig.~\ref{fig:3_paradigm} shows the progression of panoramic dental radiograph analysis from object detectors to single-pass VLMs and finally to agentic VLMs.  
Our work is the first to adopt agentic VLMs for dental panoramic analysis by introducing diagnostic tool reasoning and interactive decision-making.  
\noindent\textbf{Paradigm 1: Traditional Object Detection.}
Early systems relied on detectors such as YOLO~\cite{redmon2016you} and DETR~\cite{carion2020end} to identify lesions through bounding-box predictions:
\begin{equation}
\mathcal{D}_{\text{obj}}(I_0)
= \bigl\{(b_i,\, c_i,\, s_i)\bigr\}_{i=1}^{N},
\quad b_i \in [0,1]^4.
\end{equation}
Although these models can localize lesions accurately, their outputs are limited to box coordinates and confidence scores, leaving the diagnostic rationale unexplained and restricting the system to perception-driven inference.

\noindent\textbf{Paradigm 2: Single Forward VLMs.}
Single-pass VLMs, represented by LLaVA~\cite{liu2023visual}, generate responses directly from an image $I_0$ and a query $q$:
\begin{equation}
y = \mathcal{G}_{\text{vlm}}(I_0, q)
= \operatorname{Dec}\!\bigl(f_{\theta}(I_0),\, q\bigr).
\end{equation}
Here, $f_{\theta}(I_0)$ extracts visual features, and $\operatorname{Dec}(\cdot)$ produces the diagnostic text.  
This paradigm provides semantic interpretation but remains static, as the model responds in one pass without revisiting ambiguous regions and refining its answer.

\noindent\textbf{Paradigm 3: Agentic VLMs.}
Our OralGPT-Plus introduces an iterative thought–action–observation loop that emulates clinical diagnostic procedures.
At each step $i$, the policy produces a thought $T_i$, selects an action $\mathcal{A}_i$, and obtains a corresponding observation $O_i$:
\begin{equation}
\begin{aligned}
T_i &= \pi_{\text{thought}}(q, I_0, \mathcal{H}_{i-1}), \\
\mathcal{A}_i &= \pi_{\text{action}}(T_i), \\
O_i &= \operatorname{Env}(\mathcal{A}_i; I_0), \\
\mathcal{H}_i &= \mathcal{H}_{i-1} \cup \{T_i, \mathcal{A}_i, O_i\}.
\end{aligned}
\end{equation}
The environment $\operatorname{Env}(\cdot)$ executes image actions such as ``Zoom-In'' for close examination.
The process ends when the policy issues a Finalize action or reaches $K$ steps, producing the final report:
\begin{equation}
y^* = \pi_{\text{answer}}(q, I_0, \mathcal{H}_K).
\end{equation}
Through iterative observation and tool-guided inspection, the model learns to reassess uncertain areas, perform bilateral comparisons, and articulate clinically interpretable conclusions.

The transition from detectors to agentic VLMs reflects a shift from static perception to interactive reasoning.  
Traditional detectors localize abnormalities, single-pass VLMs describe them, and agentic VLMs analyze and verify through tool-guided, multi-step exploration.  
This progression enhances interpretability and strengthens clinical alignment in panoramic dental radiograph analysis.

%% file: 3_methods.tex
\begin{figure*}[!ht]
  \centering
  \includegraphics[width=\textwidth]{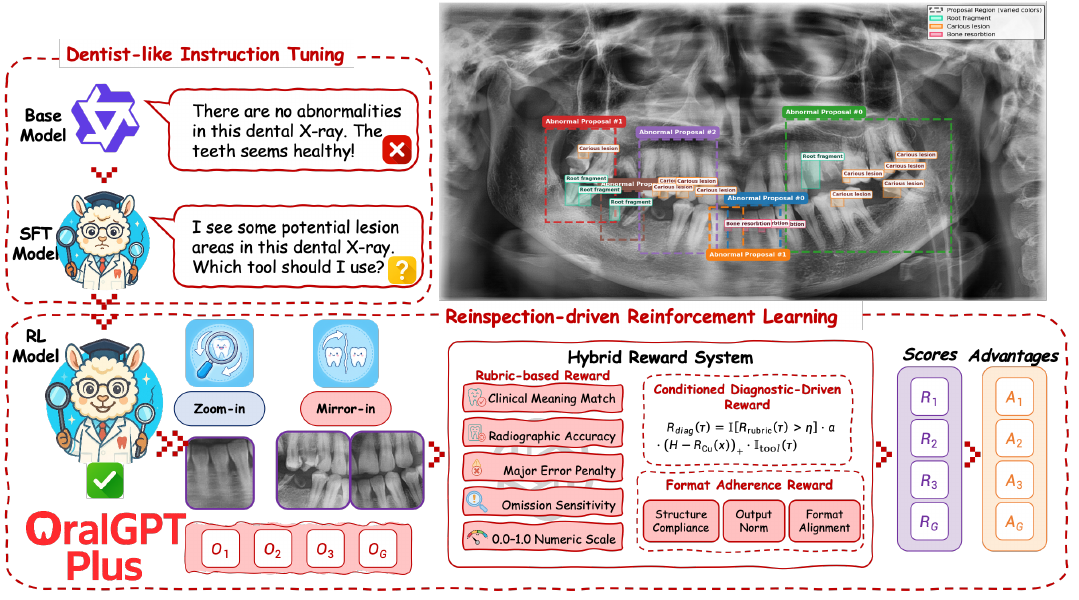}
  \caption{\textbf{Our training pipeline from dentist-like instruction tuning to re-inspection reinforcement learning.}}
  \label{fig:method}
    \vspace{-0.4cm}
\end{figure*}

\section{OralGPT-Plus}
\label{sec:methods}
In this section, we introduce OralGPT-Plus, which emulates dentists’ diagnostic reasoning on panoramic X-rays through tool-augmented learning. We first present Dentist-like Instruction Tuning~\ref{sec:sec:dentist_like_sft}, where the model acquires essential diagnostic behaviors using curated expert trajectories from the DentalProbe dataset. We then describe Reinspection-Driven Reinforcement Learning~\ref{sec:sec:reinspection_drive_rl}, which incorporates rubric-based continuous rewards and conditioned diagnostic-driven incentives to encourage clinically meaningful reinspection. These components are unified through a Hybrid Reward System that stabilizes long-horizon optimization. Together, they enable OralGPT-Plus to follow a dentist-like workflow and progressively strengthen its diagnostic competence.

\subsection{Dentist-like Instruction Tuning}
\label{sec:sec:dentist_like_sft}
\begin{figure*}[!ht]
  \centering
  \includegraphics[width=\textwidth]{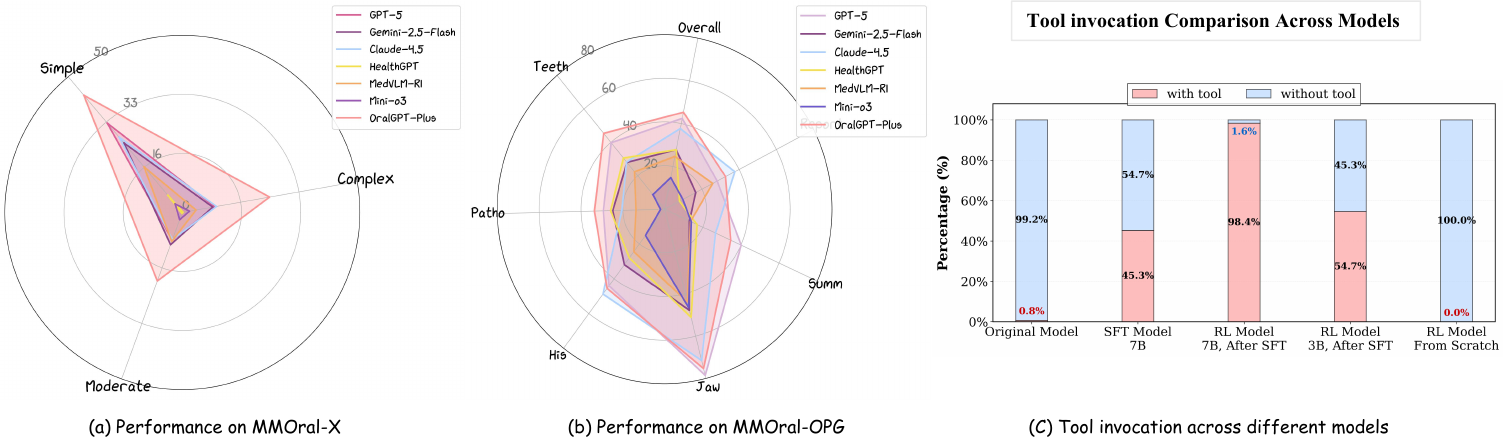}
  \caption{\textbf{Distribution across different task of different models and tool invocation comparison.}(a) The performance of the MMOral-X benchmark. (b) Performance comparison on MMOral-Omni benchmark. (c) Performance comparison on MMOral-OPG benchmark. 
  }
\vspace{-0.4cm}
  \label{fig:fig_4_performance}
\end{figure*}

Our goal is to enable VLMs to emulate dentists’ diagnostic reasoning on dental X-ray images, particularly in regions with subtle lesions. The model uses diagnostic tools to locate, zoom in, and compare suspicious areas, thereby improving panoramic X-ray analysis. However, as shown in Fig.~\ref{fig:fig_4_performance}, models such as Qwen2.5-VL fail to develop tool-based reasoning through prompt engineering alone, consistent with prior observations~\cite{wang2025pixel}. This difficulty stems from the absence of tool-interaction data during pretraining. To overcome this issue, in Sec.~\ref{sec:sec:sec:data_curation} we introduce a precise diagnostic data construction process that produces high-quality expert trajectories in Sec.~\ref{sec:sec:tool_design} and Sec.~\ref{sec:sec:dentalprobe_construction}. 

\subsubsection{Data Statistics}
\label{sec:sec:sec:data_curation}
We integrate four public datasets~\cite{panetta2021tufts, zhang2023children, hamamci2023dentex, Mureanu2024AutomatingDC} for abnormality detection in panoramic radiographs, each offering region-level annotations of dental abnormalities. These datasets cover over 50 disease categories across geographic regions, including the United States, China, Switzerland, and Romania, which enhances representativeness and mitigates sampling bias. In addition, we select 2,500 high-quality radiographs from the MMOral-OPG~\cite{hao2025towards} dataset and combine them with 2,562 collected samples to construct the 5k-image DentalProbe dataset. Detailed information about data sources and disease categories are provided in the \textit{Appendix}.

\begin{figure}[t]
  \centering
  \includegraphics[width=0.48\textwidth]{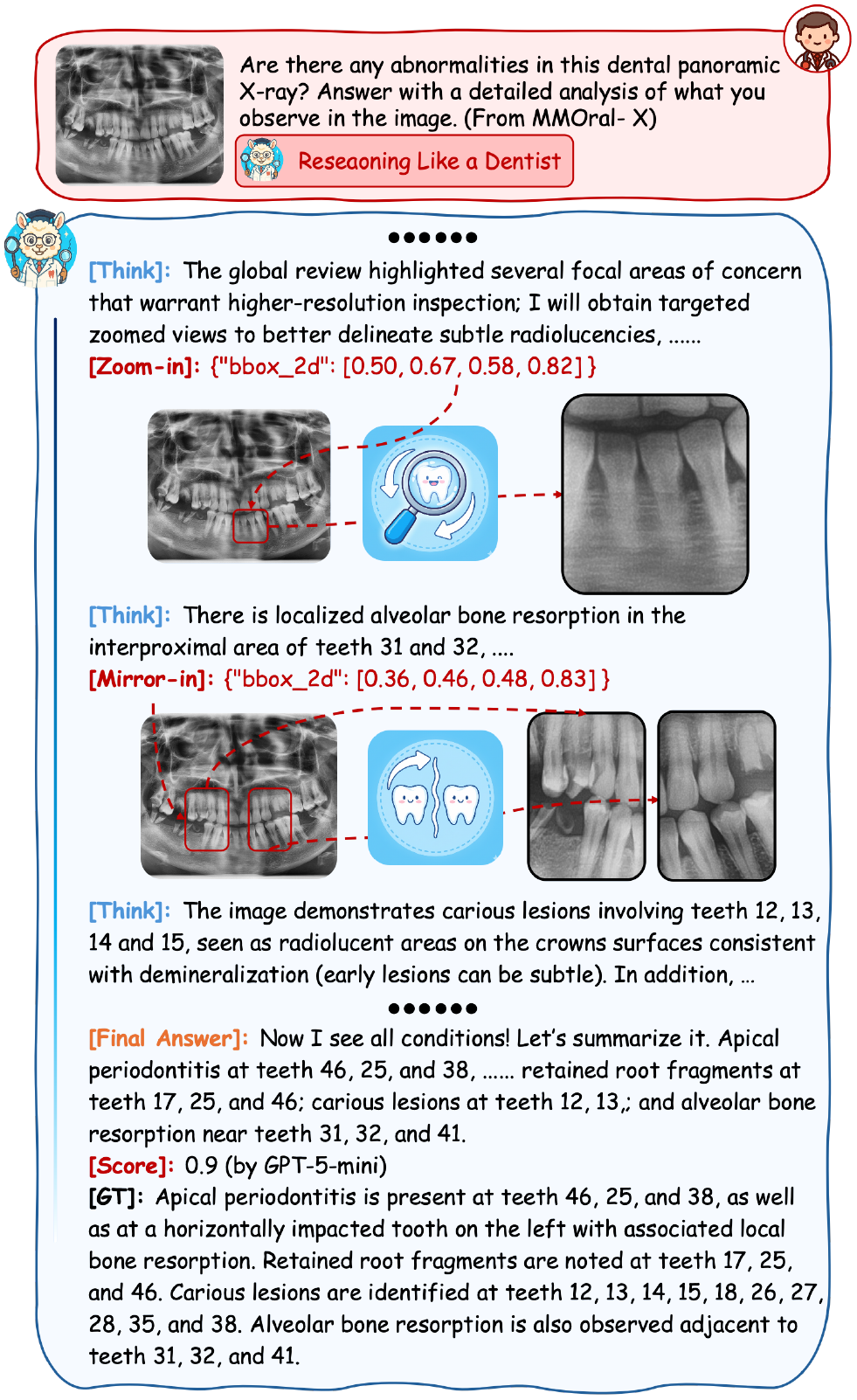}
  \caption{\textbf{Diagnostic trajectory generated by OralGPT-Plus on a sample of MMOral-X}. The model produce a final panoramic-level diagnosis and got a score 0.9.}
  \vspace{-0.4cm}
  \label{fig:case}
\end{figure}

\subsubsection{Dental-Aware Tool Design}
\label{sec:sec:tool_design}
Previous agentic VLMs rely primarily on the ``Zoom-In'' operation to magnify suspicious regions for detailed inspection. However, panoramic dental X-rays exhibit strong anatomical symmetry, and a purely local enhancement strategy often overlooks this intrinsic characteristic. In clinical practice, dentists routinely perform contralateral comparisons by examining the corresponding tooth or region on the opposite side, which allows them to determine whether subtle shadows or irregularities indicate true lesions.

Motivated by this diagnostic behavior, we introduce the ``Mirror-In'' tool, which converts oral symmetry from an implicit prior into an explicit model action. After the model identifies a potential lesion, the tool retrieves its horizontally mirrored counterpart across the midline, thereby forming a dual-view pair for comparative reasoning. Formally, given an image $I(x,y)$ of width $W$ and a selected region $[x_1,x_2]\!\times\![y_1,y_2]$, the symmetric view is defined as
\begin{equation}
\begin{aligned}
I_{\text{mirror}}(x,y) &= I(W - x,\, y),\\
(x,y) &\in [x_1,x_2]\times [y_1,y_2].
\end{aligned}
\end{equation}
In implementation, we use $1.5\times$ padded original/mirrored crops to absorb minor misalignment and asymmetry. This explicit symmetry-based inspection enhances the model’s spatial understanding and facilitates the verification of subtle or low-contrast abnormalities through contralateral reference. By embedding clinical comparison habits into the reasoning loop, the agent produces more stable and interpretable diagnostic judgments.

\subsubsection{DentalProbe Trajectory Construction}
\label{sec:sec:dentalprobe_construction}
To construct dentist-like diagnostic trajectories, we design a multi-stage pipeline that emulates clinical reasoning on panoramic X-rays as shown in Fig.~\ref{fig:data_curation}. For each OPG, a tooth-number detection model localizes tooth boxes and aligns them with lesion labels. Lesions are categorized into obvious, subtle, and bone-based findings. For subtle and bone-based cases, we apply $k$-means clustering to tooth-level boxes to generate region proposals that highlight diagnostically challenging areas requiring focused inspection.

We then construct multi-turn diagnostic trajectories by applying a set of explicit rules derived from the proposals. The initial query performs a global inspection, and subsequent turns progressively incorporate zoomed-in regions and contextual cues to mirror dentists’ stepwise reasoning. After generating these rule-based trajectories, we conduct a dedicated quality refinement stage. A multi-agent tool decision module examines each step and verifies whether the invoked tool is appropriate. If the judge agent determines that the current view is sufficient, the system validates the ``Zoom-In'' decision; otherwise, it triggers ``Mirror-In'' to obtain a symmetric comparison that supports more reliable diagnostics. Next, the visual description agent produces region-level visual summaries for all tool-invoked images. These are combined with curated Wikipedia-style clinical definitions to rewrite each reasoning step. To enhance linguistic diversity and reduce stylistic bias, we further employ three models~\cite{gpt5,claude4_5,team2023gemini} to perform independent rewrites and select the best refined version. This process yields trajectories that are clinically aligned, stylistically diverse, and instruction-ready. We finally invited expert dentist to do sampling-based evaluation on DentalProbe dataset, the score is shown in Fig.~\ref{fig:dentist_evaluation}. Through dentists’ evaluations and iterative refinement, we further improved the data quality, ensuring it meets a high standard.
We then construct multi-turn diagnostic trajectories by applying a set of explicit rules derived from the proposals. The initial query performs a global inspection, and subsequent turns progressively incorporate zoomed-in regions and contextual cues to mirror dentists’ stepwise reasoning. After generating these rule-based trajectories, we conduct a dedicated quality refinement stage. A multi-agent tool decision module examines each step and verifies whether the invoked tool is appropriate. If the judge agent determines that the current view is sufficient, the system validates the ``Zoom-In'' decision; otherwise, it triggers ``Mirror-In'' to obtain a symmetric comparison that supports more reliable diagnostics. When symmetry quality is unreliable (e.g., severe asymmetry or low-quality mirrored evidence), the judge agent routes the step back to alternative local inspection (e.g., ``Zoom-In'') rather than forcing mirrored comparison. Next, the visual description agent produces region-level visual summaries for all tool-invoked images. These are combined with curated Wikipedia-style clinical definitions to rewrite each reasoning step. To enhance linguistic diversity and reduce stylistic bias, we further employ three models~\cite{gpt5,claude4_5,team2023gemini} to perform independent rewrites and select the best refined version. This process yields trajectories that are clinically aligned, stylistically diverse, and instruction-ready. We finally invited expert dentist to do sampling-based evaluation on DentalProbe dataset, the score is shown in Fig.~\ref{fig:dentist_evaluation}. Through dentists’ evaluations and iterative refinement, we further improved the data quality, ensuring it meets a high standard.

\subsubsection{Training Strategy}
\label{sec:sec:sec:sft_training_strategy}
We apply full-parameter SFT to the language module while freezing the vision encoder and projector, ensuring stable visual features and focusing learning on grounded reasoning and correct tool use. Given multimodal input $(x, I)$ and the curated trajectory $y^*$, the training objective is:
\begin{equation}
\mathcal{L}_{\text{SFT}}
= -\sum_{t=1}^T \log \pi_\theta\!\left(y_t^* \mid x, I, y_{<t}^*\right).
\end{equation}

\input{main_performance}
\subsection{Reinspection-Driven Reinforcement Learning}
\label{sec:sec:reinspection_drive_rl}
After SFT, the model can invoke basic tools, but naive GRPO~\cite{shao2024deepseekmath,tao2025moss} remains unreliable for clinical diagnosis. Panoramic radiographs contain multiple spatially distributed lesions, while binary ${0,1}$ rewards offer all-or-nothing feedback, causing sparse signals, weak advantages, and unstable learning. Uncontrolled exploration further drives the agent toward clinically irrelevant regions.

To mitigate reward sparsity and align exploration with diagnostic reasoning, we introduce three complementary components. The rubrics-based reward provides continuous, case-aware supervision; the Conditioned Diagnostic Driven Reward activates exploration only under sufficient rubric confidence; and a unified hybrid reward integrates these signals with format adherence to ensure stable and interpretable RL.

\subsubsection{Rubrics-based Reward}
\label{sec:rubrics_scoring}
To address reward sparsity in multi-lesion scenarios, we introduce a rubric-based dental reward that provides continuous and informative supervision. Instead of using a binary $\{0,1\}$ score, we employ a few-shot rubric evaluator aligned with MMOral-X criteria to assess clinical meaning, radiographic accuracy, major error severity, and omission sensitivity. This approach ensures that partially correct predictions, such as identifying some but not all lesions, receive proportional rewards rather than collapsing to zero, thereby enabling stable and effective gradient updates.

Formally, the rubric evaluator (GPT-5-mini) assigns each trajectory $\tau$ a continuous score $R_{\text{rubrics}}(\tau)\!\in\![0,1]$. The scoring follows explicit clinical rules: synonymous diagnostic phrasing is accepted, clinically consistent tooth-indexing systems are treated as equivalent, minor coordinate omissions incur small penalties, and disease mismatches or false assertions are penalized as major errors. These structured criteria bridge the gap between binary correctness and continuous clinical judgment, offering dense supervision that rewards clinically meaningful outputs while discouraging hallucinations and omissions of key findings. This mechanism improves training efficiency and supports the conditioned diagnostic-driven reward described next.

\subsubsection{Conditioned Diagnostic-Driven Reward}
\label{sec:cond_diagnostic_reward}

Inspired by the clinical observation that dentists reinspect only when the initial diagnosis is reliable yet incomplete, we design a confidence-conditioned exploration reward that prevents unnecessary visual operations at low diagnostic certainty. We adopt the curiosity bonus from ~\cite{wang2025pixel}, enabling it only when the rubric score indicates sufficient diagnostic reliability.

For a generated trajectory $\tau$, let $R_{\text{rubrics}}(\tau)\!\in\![0,1]$ denote the diagnostic confidence evaluated by the rubric-based assessor, and let $\mathrm{Cu}(x)$ represent the current exploration saturation level for query $x$. We further define $\mathbb{1}_{\mathrm{tool}}(\tau)\in\{0,1\}$ as an indicator of whether tool is used. The conditioned diagnostic-driven reward is then formulated as:
\begin{equation}
\begin{aligned}
R_{\text{diag}}(\tau) =\;&
\mathbb{I}\!\big(R_{\text{rubrics}}(\tau) > \eta\big)\,
\cdot \alpha \\[3pt]
&\cdot (H - \mathrm{Cu}(x))_{+}\,
\cdot \mathbb{1}_{\mathrm{tool}}(\tau),
\end{aligned}
\label{eq:conditioned_curiosity}
\end{equation}
where $(\cdot)_{+}=\max(\cdot,0)$ ensures that the curiosity magnitude decreases as exploration becomes saturated, and $\alpha$ controls the strength of the intrinsic reward.

These ensure that the reward encourages clinically motivated reinspection rather than indiscriminate tool use, allowing visual operations only when they meaningfully refine an already credible diagnostic hypothesis.

\subsubsection{Hybrid Reward System}
\label{sec:hybrid_reward}

Building on the above reward components, we combine the rubric-based clinical reward, conditioned diagnostic incentive, and format signals into a unified Hybrid Reward System:
\begin{equation}
R(\tau) = R_{\text{rubrics}}(\tau) + R_{\text{format}}(\tau) + R_{\text{diag}}(\tau).
\end{equation}
This composite reward balances diagnostic accuracy, structural correctness, and exploration efficiency. It guides the agent from reliable local assessments to clinically meaningful reinspections while preserving adherence to the diagnostic schema, thereby stabilizing training and preventing tool-use collapse during long-horizon reasoning.

\subsubsection{Optimization Objective}
\label{sec:optimization_obj}

We train the model under the GRPO objective~\cite{shao2024deepseekmath} with clipped ratios, using standardized advantages denoted as $\mathbb{A}_i$ to avoid notational conflict with earlier action symbols:
\begin{align}
\mathcal{J}_{\mathrm{GRPO}}(\theta)
&= \mathbb{E}_{\,q\sim\mathcal{D},\,o_i\sim\pi_{\theta_{\text{old}}}}
   \Biggl[\frac{1}{G}\sum_{i=1}^G
   \min\!\bigl(r_i\,\mathbb{A}_i,\,\hat r_i\,\mathbb{A}_i\bigr)\Biggr].
   \label{eq:grpo_obj}
\end{align}
Here, $G$ is the number of sampled rollouts per query.  
The hybrid reward $R(\tau)$ defined above populates the return used for computing advantages.

%% file: main_performance.tex
\begin{table*}[t]
    \centering
    \caption{Model performance on \textbf{MMOral-X} and open-ended VQA for \textbf{MMOral-OPG}. Results on MMOral-X are averaged over five runs, whereas MMOral-OPG is evaluated once. In each column, the best score is \textbf{bold} and the second best is \underline{underlined}.}
    \label{table:main_result}
    \tabcolsep=0.25cm
    \resizebox{\textwidth}{!}{%
    {
        \begin{threeparttable}
        \begin{tabular}{l|c|ccc|ccccccc}
            \toprule
            \multirow{2}{*}{Model}
            & \multirow{2}{*}{Params}
            & \multicolumn{3}{c|}{MMOral-X} 
            & \multicolumn{7}{c}{MMOral-OPG Open-ended VQA} \\
            \specialrule{0em}{0pt}{1pt}
            \cline{3-12}
            \specialrule{0em}{0pt}{1pt}
            & & Simple & Moderate & Complex 
            & Teeth & Patho & His & Jaw & Summ & Report & Overall \\
            \hline
            \rowcolor{mylightred} \multicolumn{12}{l}{\textit{Proprietary VLMs (API)}} \\ 
            \hline
            GPT-5~\cite{gpt5} & N/A & 32.90 & 8.30 & 9.30 & \underline{39.84} & 29.42 & 44.08 & \textbf{77.38} & \underline{40.06} & 28.16 & \underline{42.34} \\
            Gemini-2.5-Flash~\cite{team2023gemini} & N/A & 25.56 & 9.70 & 8.70 & 28.21 & 24.64 & 31.95 & 47.74 & 13.05 & 16.73 & 27.89 \\
            Doubao-seed-1-6\tnote{a}~\cite{guo2025seed1} & N/A & 20.10 & 5.80 & 5.20 & 27.19 & 18.18 & 41.72 & 59.44 & 22.38 & \underline{40.20} & 34.20 \\
            Claude-sonnet-4-5\tnote{b}~\cite{claude4_5} & N/A & 27.72 & 8.14 & 9.78 & 28.29 & 20.45 & \textbf{48.83} & 71.56 & 26.67 & 37.70 & 37.68 \\
            Qwen3-VL-235b-a22b-thinking~\cite{Yang2025Qwen3TR} & 235B & 10.24 & 3.88 & 3.42 & 27.42 & 26.36 & 39.34 & 58.50 & \textbf{40.48} & 37.70 & 35.78 \\
            Grok-4~\cite{grok-4} & N/A & 15.18 & 7.96 & 6.12 & 22.48 & 19.08 & 32.82 & 54.81 & 25.36 & 38.30 & 30.97 \\
            GLM-4.5v~\cite{Hong2025GLM45VAG} & 106B & 15.68 & 5.46 & 4.04 & 27.53 & 22.20 & \underline{46.20} & 59.56 & 22.14 & \textbf{42.70} & 35.95 \\
            \hline
            \rowcolor{mylightred} \multicolumn{12}{l}{\textit{Medical Specific VLMs}} \\ 
            \hline
            LLaVA-Med~\cite{Li2023LLaVAMedTA} & 7B & 0.20 & 0.16 & 0.10 & 0.92 & 1.52 & 0.00 & 0.00 & 0.00 & 24.50 & 4.76 \\
            MedDr~\cite{he2024meddr} & 40B & 6.14 & 4.26 & 3.08 & 22.99 & \underline{32.58} & 29.57 & 52.44 & 20.95 & 8.70 & 26.20 \\
            HuatuoGPT-V~\cite{chen2024huatuogpt} & 34B & 6.40 & 2.62 & 4.02 & 35.18 & 24.92 & 36.32 & 65.69 & 25.48 & 24.80 & 36.02 \\
            HealthGPT~\cite{lin2025healthgpt} & 32B & 6.34 & 0.34 & 0.26 & 30.64 & 25.83 & 27.98 & 51.12 & 17.02 & 8.00 & 27.80 \\
            MedVLM-R1~\cite{pan2025medvlm} & 2B & 16.80 & 8.70 & 3.92 & 22.42 & 13.71 & 24.42 & 43.88 & 13.57 & 25.80 & 24.70 \\
            MedGemma-4B~\cite{Sellergren2025MedGemmaTR} & 4B & 8.48 & 1.06 & 3.72 & 29.69 & 11.26 & 28.12 & 41.88 & 22.62 & 39.40 & 30.45 \\
            \hline
            \rowcolor{mylightred} \multicolumn{12}{l}{\textit{Open-source VLMs}} \\ 
            \hline
            LLaVA-OneVision~\cite{Li2024LLaVAOneVisionEV} & 8B & 8.66 & 3.62 & 6.08 & 28.79 & 13.64 & 30.12 & 55.12 & 20.36 & 23.20 & 29.43 \\
            Qwen2.5-VL-3B-Instruct~\cite{Bai2025Qwen25VLTR} & 3B & 1.82 & 4.96 & 3.40 & 32.01 & 20.91 & 27.55 & 54.50 & 30.12 & 5.70 & 28.79 \\
            Qwen2.5-VL-7B-Instruct~\cite{Bai2025Qwen25VLTR} & 7B & 7.22 & 4.58 & 3.70 & 17.01 & 16.10 & 11.18 & 29.41 & 9.07 & 8.20 & 15.92 \\
            \hline
            \rowcolor{mylightred} \multicolumn{12}{l}{\textit{Agentic VLMs}} \\ 
            \hline
            Mini-o3~\cite{Lai2025Minio3SU} & 7B & 3.14 & 2.18 & 2.04 & 8.72 & 1.97 & 15.15 & 46.62 & 13.93 & 9.80 & 14.72 \\
            \hline
            \rowcolor{mylightred} \multicolumn{12}{l}{\textit{Our Models}} \\ 
            \hline
            OralGPT-Plus-3B-SFT & 3B & 16.45 & 8.32 & 14.18 & 22.17 & 13.71 & 16.44 & 35.25 & 9.88 & 19.80 & 21.02 \\
            OralGPT-Plus-7B-SFT & 7B & 23.06 & \underline{11.36} & \underline{14.40} & 35.02 & 19.32 & 29.02 & 62.37 & 20.12 & 11.90 & 31.44 \\
            OralGPT-Plus-3B & 3B & \underline{33.12} & 11.24 & 10.70 & 21.58 & 17.35 & 21.29 & 35.69 & 16.07 & 31.70 & 24.48 \\
            OralGPT-Plus-7B  & 7B & \textbf{43.16} & \textbf{20.60} & \textbf{24.96} & \textbf{45.34} & \textbf{33.56} & 45.58 & \underline{75.38} & 34.64 & 32.70 & \textbf{45.35} \\
            \bottomrule
        \end{tabular}
        \begin{tablenotes}[flushleft]
            \footnotesize
            \item[a] Evaluated API snapshot dated 2025-10-15.
            \item[b] Evaluated API snapshot dated 2025-09-29.
        \end{tablenotes}
        \end{threeparttable}
    }
    }
    \vspace{-0.4cm}
\end{table*}

%% file: 4_benchmark.tex
\section{MMOral-X Benchmark}
\label{sec:mmoral_x_bench}
\subsection{Benchmark Composition}
We introduce MMOral-X, the first benchmark that enables comprehensive diagnosis from a single panoramic radiograph. It contains 300 open-ended Q\&A pairs and 686 bounding boxes covering diverse pathological findings, including disease classification, grounding, tooth-level assessments, regional evaluations, and abnormality flags. It also includes clinically relevant non-pathological findings—such as orthodontic appliances, restorations, and surgical devices—that are essential for a complete diagnosis but are largely overlooked in prior datasets. Unlike MMOral-OPG~\cite{hao2025towards}, which targets localized, lesion-specific queries, MMOral-X supports holistic diagnostic evaluation. A sample is provided in Fig.\ref{fig:case}.

To assess varying difficulty, MMOral-X is divided into three subsets—Simple, Moderate, and Complex—each containing 100 images. Simple cases require one to two reasoning steps and contain a single proposal; Moderate cases involve two or more proposals; Complex cases include multiple proposal types with substantial visual ambiguity. This stratified organization provides a controlled gradient of diagnostic complexity while maintaining a unified task formulation. More details are provided in the \textit{Appendix}.

\subsection{Evaluation Metrics}
Following established evaluation practices and prompts design~\cite{yu2023mmvetv1,yu2024mmvetv2,hao2025towards,zha2025aircopbench}, we construct a carefully designed few-shot prompt and employ GPT-5-mini~\cite{gpt5} as the judge for open-ended assessment. The prompt contains few-shot exemplars covering fully/partially correct and incorrect responses, enabling the judge to score each prediction on a continuous scale from 0.0 to 1.0 with a resolution of 0.1. Beyond this setup, we systematically analyze the choice of the judge, its potential scoring bias, and the variance observed across multiple test runs. We also perform consistency checks to verify that the evaluation procedure remains stable under repeated testing. These analyzes collectively demonstrate that our scoring framework is reliable and robust. Full details and extended results are provided in the \textit{Appendix}.

% \subsection{Quality Control}
% \label{sec:sec:quality_control}

%% file: 5_experiments.tex
\section{Experiments}
\label{sec:experiments}
\input{abalation}
\subsection{Experimental Setups}
\begin{figure}[t]
  \centering
  \includegraphics[width=0.8\linewidth]{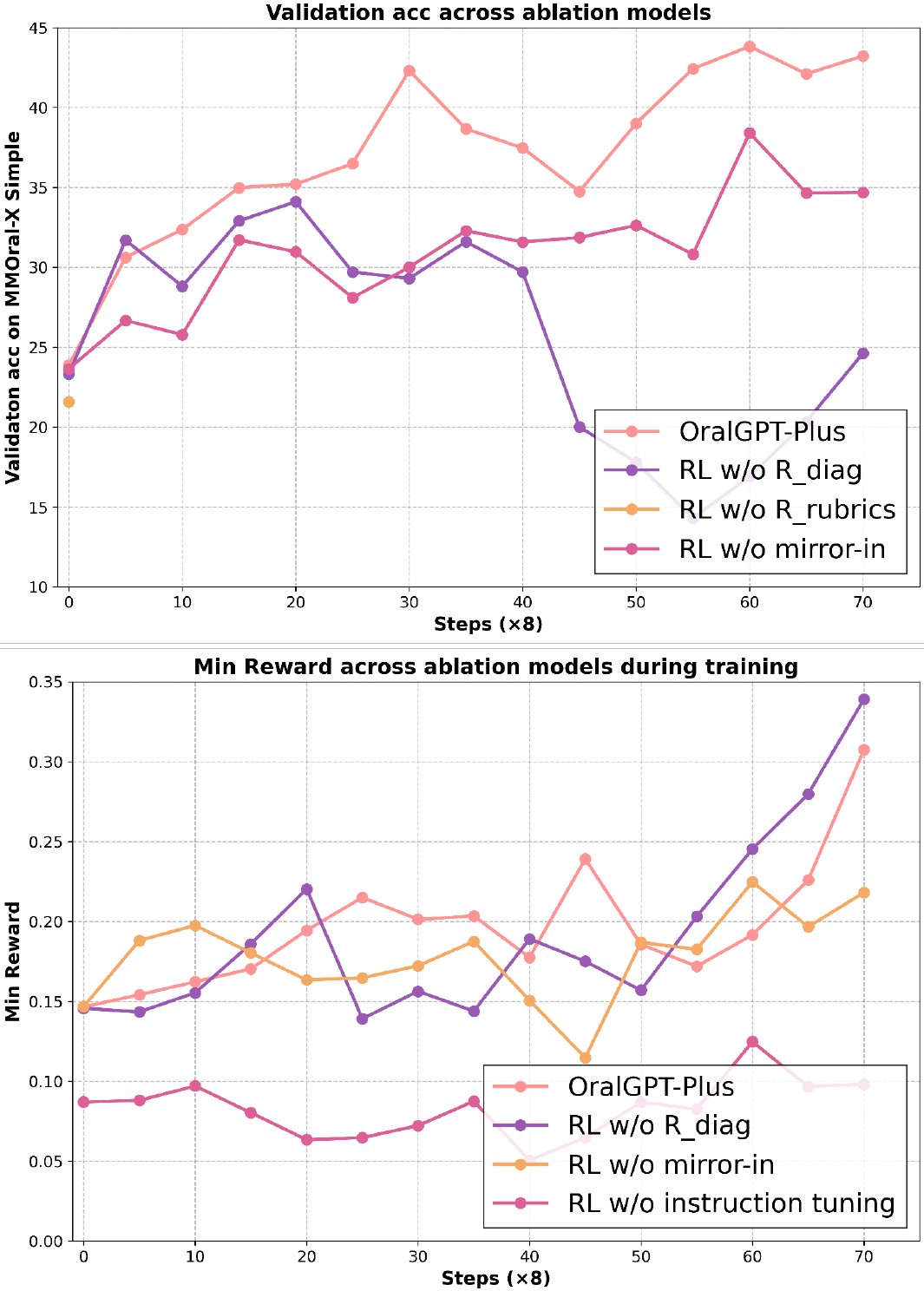}
  \caption{\textbf{Training dynamics across ablation models.}}
\vspace{-0.4cm}
  \label{fig:fig_training_dynamics}
\end{figure}
\label{sec:sec:experimental_setup}
\textbf{Settings.} All experiments are conducted on four NVIDIA A800 GPUs. For dentist-like instruction tuning, we employ Qwen2.5-VL-3B/7B-Instruct~\cite{Bai2025Qwen25VLTR}. We perform SFT on 4k DentalProbe samples for three epochs and reserve an additional 1k samples for RL to avoid data leakage. Both the rollout number and global batch size are set to 8, and all other optimization configurations follow~\cite{yu2025dapo}. At evaluation, we report $\mathrm{Avg}@5$ for MMOral-X and $\mathrm{pass}@1$ for MMOral-OPG following prior work~\cite{hao2025towards}. For judge, we chose GPT-5-mini because it offers consistent scoring, strong rubric-following, and minimal hallucination.

\noindent \textbf{Baselines.}
To enable a comprehensive comparison, we evaluate a diverse set of strong baselines
, including frontier foundation models, state-of-the-art medical multimodal models, commonly used open-source vision–language models, and tool-augmented VLMs.

\subsection{Main Results}
\label{sec:sec:main_results}
Table~\ref{table:main_result} presents the overall results of OralGPT-Plus across MMOral-X and MMOral-OPG benchmarks. On MMOral-X, OralGPT-Plus achieves the highest accuracy under all difficulty levels, surpassing powerful proprietary models such as GPT-5, as well as domain-specific models like MedDr and HuatuoGPT-V. This shows that our two-stage training, including dentist-like instruction tuning and reinspection-driven reinforcement learning, effectively enhances the model’s reasoning ability on complex dental structures. On MMOral-OPG, OralGPT-Plus also obtains the best overall score and leads in key diagnostic aspects such as Teeth, Patho, and Jaw. The results confirm that integrating structured tool interaction enables the model to perform a stepwise and clinically coherent analysis similar to that of human dentists. The distribution across multiple models on different subset are shown in Fig.~\ref{fig:fig_4_performance}.

We also note that failure cases primarily occur in complex radiographs. Additional analyses are provided in Appendix Sec.~\ref{sup:sec:training_dynamics} (tool-usage dynamics), Sec.~\ref{sup:sec:failure_analysis} (failure analysis), Fig.~\ref{sup:fig:training_dynamics}, Fig.~\ref{sup:fig:rounds_distribution}, and Table~\ref{sup:tab:stability}.

\subsection{Findings}
\label{sec:sec:in_depth_analysis}
In this section, we present four key findings derived from the results in Table~\ref{table:main_result} and the ablation studies in Table~\ref{table:ablation_study}.

\noindent\textbf{Dentist-like Instruction Tuning is Essential.}
As shown in Table~\ref{table:ablation_study} and Fig.~\ref{fig:fig_4_performance}, models without Dentist-like Instruction Tuning exhibit clear performance drops and fail to activate tool-usage behaviors, even after reinforcement learning. This demonstrates that tool-based reasoning does not emerge from reward optimization alone but requires clinically structured demonstrations that specify when and how tools should be used. In contrast, instruction-tuned models acquire a stable tool schema that enables reinforcement learning to further refine diagnostic behavior.

\noindent\textbf{Model Capacity Determines the Effectiveness of Reinforcement Learning.}
With the same instruction-tuning and reward setup, OralGPT-Plus-7B yields markedly larger RL gains than the 3B model. As shown in Fig.~\ref{fig:fig_4_performance}, the 7B model consistently develops stable tool-based reinspections, while the 3B model fails to do so. As a result, the 7B variant improves across all MMOral-X levels, indicating that sufficient model capacity is crucial for mastering tool-driven diagnostic workflows and fully benefiting from RL.

\noindent\textbf{Conditioned Diagnostic-Driven Reward Prevents Reward Hacking.}
Ablating the conditioned diagnostic-driven reward induces clear reward hacking. As shown in Fig.~\ref{fig:fig_training_dynamics} and Tab.~\ref{table:ablation_study}, the model without this component displays a collapse--rebound pattern in which tool usage first drops and then spikes while validation accuracy deteriorates. This failure arises because the agent exploits the reward by invoking a single tool and answering without meaningful inspection. Incorporating the conditioned mechanism suppresses this shortcut, stabilizes training, and triggers reinspections only when diagnostic confidence is sufficient, thereby aligning exploration with clinical reasoning.

\noindent\textbf{Mirror-In Tool Significantly Enhances Diagnostic Reliability.}
Panoramic radiographs are highly symmetric, and the ``Mirror-In'' tool effectively exploits this structure. Removing it leads to clear drops across all difficulty levels shown in Tab.\ref{table:ablation_study}: accuracy on Simple decreases from 43.22 to 34.68, on Moderate from 20.60 to 14.26, and on Complex from 24.96 to 14.30. These declines indicate the model struggles with low-contrast ambiguities and exhibits unstable reward trajectories. Integrating ``Mirror-In'' introduces symmetry-aware verification, stabilizes training, and significantly improves diagnostic robustness.

%% file: abalation.tex
\begin{table}[t]
    \centering
    \caption{Ablation studies.}
    \label{table:ablation_study}
    \tabcolsep=0.3cm
    \resizebox{\linewidth}{!}
    {
        \begin{threeparttable}
        \begin{tabular}{l|ccc}
            \toprule
            \multirow{2}{*}{Model}
            & \multicolumn{3}{c}{MMOral-X} \\
            \specialrule{0em}{0pt}{1pt}
            \cline{2-4}
            \specialrule{0em}{0pt}{1pt}
            & Simple & Moderate & Complex \\
            \midrule
            wo Instruction tuning & 8.02 & 4.64 & 4.98 \\
            wo $R_{rubrics}$ and $R_{cond}$ & 20.82 & 9.24 & 9.82 \\
            wo $R_{rubrics}$  & 21.48 & 12.08 & 13.42 \\
            wo $R_{cond}$ & 24.62 & 14.14 & \underline{15.96} \\
            wo ``Mirror-In'' tool  & \underline{34.68} & \underline{14.26} & 14.30 \\
            OralGPT-Plus &\textbf{43.16} & \textbf{20.60} & \textbf{24.96} \\
            \bottomrule
        \end{tabular}
        \end{threeparttable}
    }
    \vspace{-0.4cm}
\end{table}

%% file: 6_conclusion.tex
\section{Conclusion}
\label{sec:conclusion}
This work advances panoramic radiograph analysis from static perception to interactive diagnostic reasoning. We introduce OralGPT-Plus, an agentic model that inspects, compares, and verifies findings through clinically grounded tool use. With dentist-like instruction tuning and reinspection-driven reinforcement learning, OralGPT-Plus achieves strong performance on MMOral-X and MMOral-OPG, surpassing proprietary and medical-domain baselines. The results show that diagnostic capability emerges not only from model capacity but also from the ability to act, reflect, and re-examine. This paradigm provides a promising direction for future diagnostic VLMs.

%% file: X_suppl.tex
\setcounter{page}{1}
\onecolumn
\maketitlesupplementary
\renewcommand{\contentsname}{Appendix Contents}
\begin{center}
\large \textbf{Appendix Contents}
\end{center}

\vspace{-1em}
\setcounter{tocdepth}{2}
\tableofcontents
\section{Related Work}
\input{sup_diseases_list}
\label{sec:related_work}
\subsection{VLMs for Dental Panoramic X-ray}
\label{sec:related_work:oral_radiology}
Vision–language models (VLMs) have shown strong potential in oral radiology by unifying image understanding, dental knowledge, and diagnostic report generation. However, they face persistent challenges including limited domain-specific data, scarcity of fine-grained tooth-level annotations, factual reliability, and privacy constraints~\cite{huang2023chatgpt}.
Recent dental VLMs~\cite{meng2025dentvlm} typically adopt a two-stage pipeline: (i) perform vision–language alignment or basic VQA-style modeling, and (ii) apply instruction tuning with rationales to enhance open-ended descriptions and clinical coherence~\cite{zhang2025oralgpt}. Parallel to these efforts, several domain-specific datasets and benchmarks have emerged. OralGPT~\cite{hao2025towards} introduces a five-dimension evaluation showing that both general and medical VLMs exhibit large performance gaps on tooth-level and region-grounded diagnostic tasks, but that in-domain instruction tuning can markedly improve accuracy. DentalBench~\cite{zhu2025dentalbench} further provides the first comprehensive bilingual benchmark and corpus for dental LLM evaluation, demonstrating that domain adaptation is crucial for performing specialized tasks such as intervention classification or pathology interpretation.
Methodologically, these works often rely on structured cues—such as FDI tooth indices, treatment records (filling, crown, RCT, implant), and similarity-based pseudo-labeling—to compensate for supervision shortages and improve interpretability. Yet despite superior textual expressiveness, existing models still rely on single-pass inference and lack explicit representations of the multi-step inspection and contralateral comparison that dentists routinely perform. Unlike prior single-pass dental VLMs, OralGPT-Plus introduces the first agentic panoramic dental VLM capable of iterative diagnostic reasoning. Through curated multi-turn trajectories (DentalProbe), explicit diagnostic tools (``Zoom-In'' + ``Mirror-In''), and reinspection-driven reinforcement learning, OralGPT-Plus aligns directly with real clinical workflows rather than relying solely on static image–text alignment.

\subsection{Thinking with Images in VLMs}
\label{sec:related_work:thinking_with_images}
Recent advances in multimodal reasoning have introduced the paradigm of ``thinking with images,'' where VLMs dynamically interact with images through tools such as zooming, cropping, and region selection~\cite{gpt5, Yang2025Qwen3TR, geng2025webwatcher, Zheng2025DeepEyesI, lai2025mini, wang2025pixel}. This allows models to refine hypotheses and perform multi-step visual analysis. However, existing approaches lack medical priors, do not leverage expert multi-step trajectories, and may exhibit uncontrolled exploration, making them unsuitable for clinical diagnosis.  
Panoramic dental radiographs naturally fit this paradigm, as dentists routinely zoom into subtle regions and compare symmetric structures. Yet prior work has not adapted thinking-with-images to panoramic diagnosis. We address this gap by constructing DentalProbe for expert-guided trajectories, introducing ``Mirror-In'' for symmetry-aware inspection, and developing a reinspection-driven RL framework that aligns exploration with diagnostic reliability.

\section{Details of DentalProbe Curation}
\label{sup:sec:data_curation_detail}
\subsection{Data Statistics Details}
\label{sup:sec:sec:details}
As shown in Tab.~\ref{sup:tab:data_sources_compact}, we integrate four publicly available panoramic radiograph datasets and select part of suitable images: the Tufts Dental Database (380 images)~\cite{panetta2021tufts}, the Children’s Panoramic Dataset (69 high-quality pediatric cases)~\cite{zhang2023children}, the fully annotated DENTEX Challenge dataset (678 images)~\cite{hamamci2023dentex}, and the Diagnostics TEAM dataset (1435 images)~\cite{Mureanu2024AutomatingDC}.
Together, these sources contribute a total of 2,562 clinically annotated panoramic radiographs covering multiple demographic groups, including both children and adults, and originating from geographically diverse regions such as the United States, China, Switzerland, Turkey, and Romania. 
Across these datasets, more than 50 dental conditions are labeled, including caries, periapical lesions, impacted teeth, periodontal bone loss, prosthetic restorations, orthodontic appliances, and rare developmental anomalies (the details of diseases list are shown in Tab.~\ref{sup:tab:diseases_list_details}). 
DENTEX and Tufts provide tooth-level enumeration that follows the FDI two-digit notation system, while the pediatric dataset contributes mixed-dentition cases that reflect developmental-stage variability. 
This diverse and globally distributed composition substantially enhances the representativeness, robustness, and clinical generalizability of the final DentalProbe dataset and MMOral-X benchmark.
\input{sup_data_source}
\input{sup_rewrite_wiki_description} 

\subsection{Trajectory Construction Details.}
\label{sec:sec:trajectory_construction}
In this section, we introduce the details of how we build the trajectories of DentalProbe dataset.

To construct high-quality diagnostic trajectories, we design a multi-stage, multi-agent generation process that emulates the reasoning workflow of dentists on panoramic X-rays.

\input{sup_rule_based_traj_type}
\paragraph{Step 1: Tooth Detection and Lesion Alignment.}
For each OPG sample, a tooth-number detection expert model identifies all tooth bounding boxes and aligns them with the corresponding ground-truth radiograph findings annotations. Findings are categorized into three types shown in Tab.~\ref{sup:tab:finding_categories}: (a) obvious findings, (b) subtle findings, and (c) bone-based findings. This taxonomy guides the subsequent proposal generation and reasoning flow. 

\paragraph{Step 2: Proposal Generation.}
For subtle and bone-based findings, we apply $k$-means clustering to the detected tooth-level bounding boxes to generate multiple region proposals. Each proposal represents a potentially ambiguous or hard-to-diagnose area requiring focused analysis in later reasoning steps.

\paragraph{Step 3: Multi-Agent Trajectory Synthesis}
We then construct multi-turn reasoning trajectories under a rule-based template:
\begin{itemize}
    \item \textbf{Round 1:} The initial query $Q_1$ focuses on global inspection. The model outputs $A_1$ emphasizing obvious findings and triggers a ``Zoom-In'' tool call to the detected region.
    \item \textbf{Round 2 and onward:} The zoomed-in result becomes the next-round input $Q_{i+1}$, where the query template is extended with contextual cues (\textit{``After the above Action X...''}). The model continues reasoning and generating the corresponding answer $A_{i+1}$.
    \item \textbf{Tool Selection with Multi-Agent Collaboration:} For each zoomed-in image, answer agent performs diagnostic Q\&A, while judge agent evaluates the answerability. If the case can be correctly answered, the proposal is retained under ``Zoom-In''; otherwise, the system invokes ``Mirror-In'' to create a symmetric comparison view. visual describe agent then produces detailed visual descriptions for each step.
\end{itemize}
We employ GPT-5~\cite{gpt5} to perform different roles of this multi-agent process, the prompts are shown in Fig.~\ref{prompt1}, Fig.~\ref{prompt2}, Fig.~\ref{prompt3}, Fig.~\ref{prompt4}. This iterative process yields a complete multi-turn diagnostic trajectory incorporating both local inspection and symmetry-aware comparison.

\paragraph{Step 4: Linguistic Enhancement and Visual Alignment.}
For each Q/A pair, we refine textual reasoning by integrating two complementary sources: (i) standard radiographic descriptions retrieved from Wikipedia according to the ground-truth disease category, and (ii) visual features extracted by agent 3. The merged content is rewritten into coherent diagnostic narratives, ensuring consistency between image evidence and textual explanation. For each category, the radiographic descriptions we use can be seen in Tab.~\ref{sup:tab:categories_radiographic_desc}.

Through this multi-agent, multi-round synthesis pipeline, we obtain high-quality trajectories that encapsulate clinical reasoning patterns, structured tool usage, and visually grounded explanations—serving as the foundation for Dentist-like Instruction Tuning.

\subsection{Teeth ID Detection.}
For teeth id detection, we use the expert model from~\cite{hao2025towards} trained with 2796 samples from 3 open-sourced datasets as shown in Tab.~\ref{sup:tab:teeth_id_detection_model}, which labeled with 1 to 32 tooth numbering following the FDI tooth numbering system. The teeth detection ability is powerful and robust.
\input{sup_expert_model}

\subsection{Dental-Aware Tool Design Details} 
\label{sup:sec:sec:dental_aware_tool_design}
Conventional agentic visual reasoning models commonly rely on the ``Zoom-In'' operation to enlarge suspicious regions for detailed observation. However, such a purely local enhancement often neglects the inherent anatomical symmetry of the oral cavity. In clinical practice, dentists routinely perform ``contralateral comparison'' to analyze the corresponding tooth or region on the opposite side of the jaw—to evaluate whether a subtle finding (e.g., shadow, opacity, or margin irregularity) reflects a true lesion or merely a radiographic artifact. This diagnostic habit leverages the structural bilateral symmetry of the dentition and surrounding bone, providing a natural reference baseline for uncertain cases.

Inspired by this clinical rationale, we transform oral symmetry from an implicit prior into an explicit and callable diagnostic action through a novel tool, `Mirror-In'. The ``Mirror-In'' operation allows the model, after localizing a potential lesion, to automatically retrieve the horizontally symmetric region across the midline, thereby constructing a paired comparison view that emulates human inspection behavior. Formally, let $I(x,y)$ denote the panoramic X-ray image of width $W$. For a selected region $[x_1,x_2]\!\times\![y_1,y_2]$, the mirrored counterpart is defined as:
\begin{equation}
\begin{aligned}
I_{\text{mirror}}(x,y) &= I(W - x,\, y),\\
(x,y) &\in [x_1,x_2]\times [y_1,y_2].
\end{aligned}
\end{equation}
This operation reflects pixel coordinates along the vertical midline and returns a horizontally flipped counterpart of the selected patch, forming a structured dual-view pair $(I_{\text{crop}}, I_{\text{mirror}})$ for subsequent reasoning.

Compared with the conventional ``Zoom-in'' that focuses solely on intra-patch details, the ``Mirror-In'' tool introduces an inter-patch relational perspective. It enables the model to evaluate whether a suspected finding deviates from its symmetric counterpart, effectively providing a self-calibrated ``reference control.'' This mechanism substantially mitigates diagnostic uncertainty arising from local exposure imbalance, motion blur, or anatomical variation. For instance, small carious lesions, enamel defects, or periapical radiolucencies can be better verified by comparing with the healthy contralateral region. From a modeling standpoint, ``Mirror-In'' enriches the agent’s observation space with cross-quadrant contrastive cues, promoting relational consistency and spatial awareness beyond the local receptive field. 

Clinically, the ``Mirror-In'' design resonates with the dentist’s habitual reasoning process—``see, compare, and conclude''—while computationally it offers a structured way to embed symmetry priors into visual reasoning loops. By enabling symmetry-aware contrastive inspection, our agent achieves more stable judgments for subtle or low-contrast abnormalities and demonstrates stronger generalization under radiographic noise and patient-specific variation.
\begin{figure*}[t]
  \centering
  \includegraphics[width=0.8\linewidth]{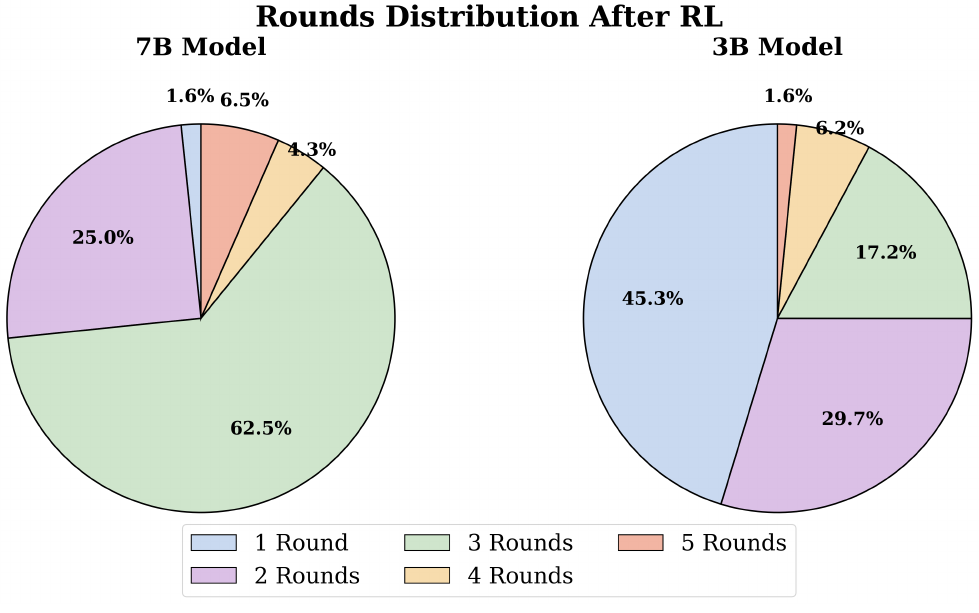}
  \caption{\textbf{Rounds distribution after RL of OralGPT-Plus 7B and 3B.}}
  \label{sup:fig:rounds_distribution}
\end{figure*}

\begin{figure*}[htbp]
  \centering
\includegraphics[width=0.8\linewidth]{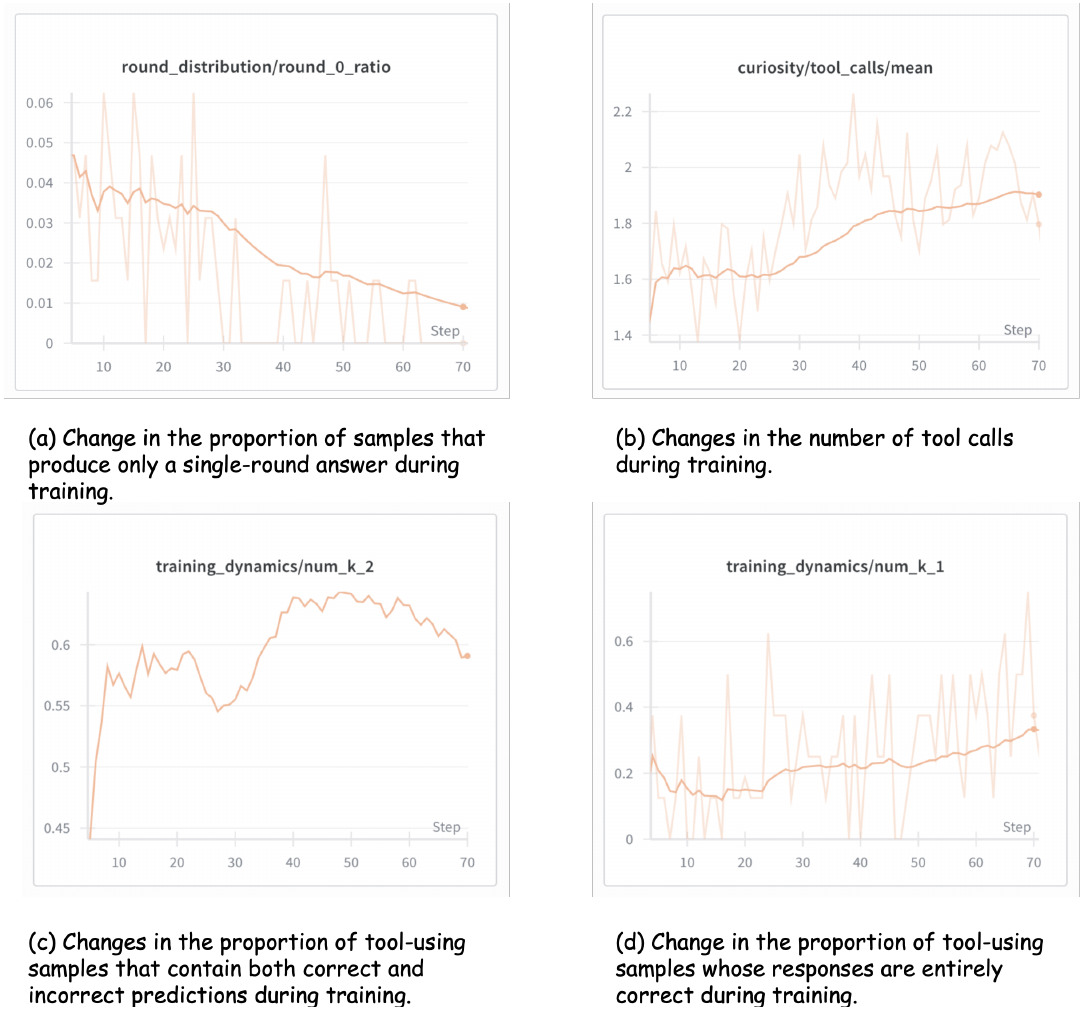}
  \caption{Evolution of key training dynamics throughout reinspection-driven reinforcement learning.} 
\label{sup:fig:training_dynamics}
\end{figure*}
\input{sup_stability_judge}
\section{Stability and Consistency for Evaluation}
In this section, we present the systemmatic analysis of the evaluation metric for MMOral-X. The analysis can be seen in ~\ref{sup:tab:stability}
Since using LLMs as judges inevitably introduces randomness, even with the temperature hyperparameter set to 0, we conduct multiple repeated experiments to verify the stability of LLMs as judges. Specifically, we evaluate the prediction results of GPT-5~\cite{gpt5}, Gemini-2.5-Flash~\cite{team2023gemini}, GLM-4.5v~\cite{Hong2025GLM45VAG}, and Qwen2.5-VL-7B-Instruct~\cite{Bai2025Qwen25VLTR} on the MMOral-X benchmark using GPT-5-mini~\cite{gpt5} with the same prompt five times. The obtained mean, standard deviation, and coefficient of variation (CV) of the metric ``overall'' are shown in Table~\ref{sup:tab:stability}. For proprietary models, medical-specific models, and general-purpose LVLMs, the standard deviation of the metric "overall" is no more than 0.212 when evaluated 5 times using GPT-5-mini with our designed few-shot prompt. Specifically, for the prediction results of GPT-5, the standard deviation of the scores is 0.179, while for Qwen3-VL-8B, the standard deviation is as low as 0.039. Meanwhile, CV (Coefficient of Variation), as a standardized measure of dispersion of a probability distribution, can be used to assess the stability of scores across multiple experiments. The CV values for the prediction results of these four models, after being scored 5 times, are all around 0.5\%, which demonstrates the evaluation stability of using LLMs as evaluators.  

\section{Training Details}
\input{sup_hyper_para}
\subsection{Hyperparameters of Training.}
The hyperparameters of dentist-like instruction tuning and Reinspection-driven reinforcement learning are shown in Tab.~\ref{sup:tab:training_details_sft} and Tab.~\ref{sup:tab:training_details_rl}. For instruction tuning, we follow the basic setting of LLaMA-Factory~\cite{zheng2024llamafactory}. For reinforcement learning, we follow the setting of Mini-o3~\cite{lai2025mini} using Verl framework ~\cite{sheng2024hybridflow}.
\subsection{Rubrics-based Reward}
For rubrics-based reward, the prompt we use for can be seen at Fig.~\ref{fig:prompt:system_prompt} and Fig.~\ref{fig:prompt:query_prompt}.

\subsection{Conditioned Diagnostic-Driven Reward}
\label{sup:sec:cond_diagnostic_reward}

Following the confidence-gated curiosity mechanism in PixelReasoner~\cite{wang2025pixel}, we design a conditioned diagnostic-driven reward that activates visual reinspection only when the textual diagnosis is already reliable but potentially incomplete. For a query $x$ and a generated trajectory $\tau$, let $R_{\text{rubrics}}(\tau)\!\in\![0,1]$ denote the rubric-based diagnostic reliability, and let $n_{\mathrm{tool}}(\tau)$ be the number of visual tool invocations in $\tau$. We define the exploration saturation level for query $x$ as the expected tool-usage intensity under the current policy
\begin{equation}
\mathrm{Cu}(x) \doteq 
\mathbb{E}_{\tau \sim \pi_{\theta}(\cdot \mid x)}
\big[n_{\mathrm{tool}}(\tau)\big]
\approx \frac{1}{T}\sum_{t=1}^{T} n_{\mathrm{tool}}(\tau_t),
\label{eq:cu_definition}
\end{equation}
where $\{\tau_t\}_{t=1}^{T}$ are recent rollouts for the same query $x$. We further define the binary indicator
$\mathbb{1}_{\mathrm{tool}}(\tau) = \mathbb{I}\big(n_{\mathrm{tool}}(\tau) > 0\big)$,
which records whether any visual tool is used in $\tau$.

Applying a Lagrangian-style confidence gating as in PixelReasoner~\cite{wang2025pixel}, we obtain the intrinsic curiosity bonus
\begin{equation}
R_{\text{diag}}(\tau)
=
\alpha\,
\mathbb{I}\!\big(R_{\text{rubrics}}(\tau)>\eta\big)\,
\big(H-\mathrm{Cu}(x)\big)_{+}\,
\mathbb{1}_{\mathrm{tool}}(\tau),
\label{eq:conditioned_curiosity_final}
\end{equation}
where $(\cdot)_{+}=\max(\cdot,0)$ ensures that the bonus decays to zero as $\mathrm{Cu}(x)$ approaches the saturation level $H$. This confidence-conditioned formulation, directly following the spirit of PixelReasoner, encourages clinically meaningful reinspection only when the diagnosis is sufficiently trustworthy and the visual exploration budget for query $x$ is not yet exhausted.

\subsection{RL Optimization Objective}
\label{sup:sec:sec:rl_optimization}

Here we provide a detailed description of the optimization objective used in our reinspection-driven reinforcement learning procedure. We optimize the policy using the GRPO~\cite{shao2024deepseekmath} objective with clipped importance ratios, following a PPO-style constrained update strategy that stabilizes reinforcement learning on long-horizon, multi-turn reasoning tasks. To avoid overloading the action notation, we use $\mathbb{A}_i$ to denote the standardized advantage associated with the $i$-th sampled response. The optimization objective is defined as:
{\small
\begin{align}
\mathcal{J}_{\mathrm{GRPO}}(\theta)
&= \mathbb{E}_{\,q\sim\mathcal{D},\,o_i\sim\pi_{\theta_{\text{old}}}}
   \Biggl[\frac{1}{G}\sum_{i=1}^G
   \min\!\bigl(r_i\,\mathbb{A}_i,\,\hat r_i\,\mathbb{A}_i\bigr)\Biggr], \label{eq:grpo_obj}\\[3pt]
&\begin{alignedat}{2}
r_i     &= \frac{\pi_\theta(o_i\!\mid\! q)}{\pi_{\theta_{\text{old}}}(o_i\!\mid\! q)},\quad &
\hat r_i&= \operatorname{clip}(r_i,\,1-\epsilon,\,1+\epsilon)
\end{alignedat}\label{eq:grpo_ratio}\\[3pt]
\mathbb{A}_i &= \frac{r_i-\bar r}{s_r},\qquad
\bar r = \tfrac{1}{G}\!\sum_{j=1}^G r_j, \label{eq:grpo_adv}\\[3pt]
s_r &= \sqrt{\tfrac{1}{G}\!\sum_{j=1}^G (r_j-\bar r)^2}. \label{eq:grpo_std}
\end{align}
}
In this formulation, each query $q$ is sampled from the training distribution $\mathcal{D}$, and for every query the old policy $\pi_{\theta_{\mathrm{old}}}$ generates $G$ candidate outputs $\{o_i\}$, which are then evaluated to produce graded rewards. The term $r_i$ is the importance ratio that measures how the new policy $\pi_\theta$ changes the likelihood of producing the same output compared with the old policy, while the clipped version $\hat r_i$ constrains the update within the trust region $[1-\epsilon,1+\epsilon]$ to avoid excessively large policy shifts. The advantage $\mathbb{A}_i$ is computed by standardizing the set of ratios $\{r_i\}$ within the same query, making $\mathbb{A}_i$ a relative measure of how much better the $i$-th sample performs compared with other candidates for that query. This intra-query normalization stabilizes optimization and prevents the model from overfitting to a small number of high-reward trajectories. The GRPO objective then takes the minimum of the unclipped and clipped surrogate terms, ensuring that beneficial updates are encouraged while overly aggressive updates are suppressed. Through this mechanism, the model can reliably propagate graded, clinically grounded rewards into stable advantages, enabling effective learning of multi-step, symmetry-aware reasoning behaviors without collapsing into degenerate tool-use patterns.

\section{Additional Analysis}
\label{sup:sec:more_analysis} 
\subsection{Analysis on rounds distribution}
After reinforcement learning, the two models exhibit distinctly different
round distributions, highlighting the impact of model capacity on
multi-turn diagnostic behavior. As shown in Fig.~\ref{sup:fig:rounds_distribution}, the 7B model predominantly adopts longer trajectories: 62.5\% of its cases use three
rounds and 25.0\% use two rounds, while only 1.6\% terminate after a
single round. This pattern indicates that the larger model consistently
conducts deeper reinspection and maintains stable tool-based reasoning
throughout the diagnostic process. In contrast, the 3B model shows a much flatter distribution. Nearly 45.3\% of its outputs stop after a single round, and only 17.2\% reach three rounds. The early-termination behavior suggests that the smaller
model struggles to sustain multi-step examination and often fails to
trigger additional clinically meaningful inspections. 

These observations are fully consistent with our earlier findings that model capacity determines the effectiveness of reinforcement learning. The   7B model reliably develops a tool-driven diagnostic workflow and benefits substantially from RL, whereas the 3B model gains much less due to its limited ability to support stable, multi-round tool usage.

\subsection{Qualitative Analysis of Tool-Usage Behaviors}
\label{sup:sec:training_dynamics}

During reinspection-driven reinforcement learning, our model exhibits clear behavioral transitions that illustrate the emergence of a structured, dentist-like diagnostic workflow.
Figure~\ref{sup:fig:training_dynamics} summarizes four complementary and key trends reflecting how tool usage, multi-round reasoning, and correctness evolve throughout training. 

First, the proportion of single-round responses steadily decreases, indicating that the model progressively abandons simplistic one-shot predictions in favor of multi-step inspection and verification. At the same time, the average number of tool calls consistently increases, suggesting a growing reliance on zooming, mirroring, and other visual operations as the model learns to actively explore ambiguous regions rather than passively interpreting the raw image.

The fraction of tool-using samples that contain both correct and incorrect predictions initially rises and later stabilizes, reflecting a natural exploration--to--convergence trajectory. Early in training, the model experiments with diverse tool-invoked reasoning patterns; as learning progresses, this exploratory variability gradually becomes more structured. Finally, the proportion of tool-using samples whose outputs are entirely correct grows steadily, demonstrating that the model not only uses tools more frequently but also uses them more effectively. This trend confirms that the conditioned diagnostic-driven reward successfully guides the agent toward clinically meaningful reinspection behaviors and more reliable diagnostic conclusions.

Overall, these dynamics reveal a consistent shift from passive image understanding to structured, tool-mediated multi-round reasoning, which is highly consistent with the observed performance improvements.

\subsection{Scalability Analysis}
\label{sec:scalability_appendix}

The architectural design of OralGPT-Plus naturally supports scalability across model capacity, tool kinds, and reinforcement learning complexity. Larger models demonstrate stronger gains from the proposed training pipeline, as capacity directly enhances the stability of multi-turn reasoning and the reliability of tool usage. This indicates that the framework can continue to improve when scaled to models with higher parameter counts, where richer intermediate representations and more consistent inspection policies are likely to emerge.

In addition, both the tool interface and the reinforcement learning framework are constructed in a modular manner. The current thought–action–observation loop remains compatible with additional inspection tools such as multi-scale visual operators, contrast manipulation, or region highlighting, without requiring changes to the core architecture. The hybrid reward structure also supports further extension, since each reward term is defined independently and can incorporate new clinical objectives as task complexity increases. Finally, although OralGPT-Plus is developed for panoramic radiographs, the underlying reasoning paradigm generalizes to other clinical imaging domains that require localized examination and iterative verification, such as mammography, fundus imaging, CT interpretations, or ultrasound analysis. These properties collectively show that the system is well suited for scaling toward broader and more advanced medical agent applications.

\subsection{Failure Analysis}
\label{sup:sec:failure_analysis}
Failure cases mainly arise in highly complex radiographs where multiple subtle, overlapping, or low-contrast lesions appear simultaneously. In such settings, even well-timed tool usage cannot fully resolve visual ambiguity: magnified views may emphasize noise or anatomical variations, and symmetry-based comparisons may surface misleading cues. As a result, the model may over-interpret minor artifacts, omit clinically important findings, or generate confident yet incorrect explanations. These observations indicate that when a single image contains dense and visually entangled abnormalities, tool-augmented reasoning may still be insufficient and can occasionally amplify hallucinations rather than suppress them.

\section{Additional Cases}
\textbf{Proposal Cases.} Additional proposals we clustered are shown in Fig.~\ref{sup:fig:proposal_cases1}, Fig.~\ref{sup:fig:proposal_cases2}, Fig.~\ref{sup:fig:proposal_cases3} and Fig.~\ref{sup:fig:proposal_cases4}.

\noindent \textbf{DentalProbe Cases.} Additional trajectories cases of DentalProbe are shown in Fig.~\ref{sup:fig:training_sample_1}, Fig.~\ref{sup:fig:training_sample_2} and Fig.~\ref{sup:fig:training_sample_3}

\noindent \textbf{MMOral-X Cases.} Additional MMOral-X cases and comparison between the response of OralGPT-Plus and other models are shown in Fig.~\ref{sup:fig:mmoral-x-case1} and Fig.~\ref{sup:fig:mmoral-x-case2}.

\input{figures/sup_proposal_cases}
\input{figures/sup_training_samples}
\input{figures/sup_prompt}
\input{figures/sup_prompt_datagen}

%% file: sup_diseases_list.tex
\begin{table*}[htbp]
\centering
\caption{Categories of radiograph findings and abnormal conditions.}
\label{sup:tab:diseases_list_details}
\begin{tabular}{|p{5cm}|p{5cm}|p{5cm}|}
\hline
\multicolumn{3}{|c|}{\textbf{Some categories of radiograph findings and abnormal conditions}} \\ 
\hline
Apical Periodontitis & Periapical Abscess & Periapical Granuloma \\ \hline
Apical Sclerosing Osteitis & Apical Rarefying and Sclerosing Osteitis & Focal Osteomyelitis \\ \hline
Pericoronitis & Periodontal Disease / Bone Loss & Endo-Perio Combination Lesion \\ \hline
Implantitis / Peri-implantitis & Mucous Retention Pseudocyst & Osteoarthritis of the Condyle \\ \hline
Tonsilloliths & Calcified Carotid Atheromatous Plaque & Antrolith \\ \hline
Foreign Body Reaction & Oroantral Communication & Trauma from Occlusion \\ \hline
Dentigerous Cyst & Odontogenic Keratocyst (OKC) & Radicular Cyst \\ \hline
Incisive Canal Cyst & Ameloblastoma & Odontoma \\ \hline
Osteoma & Malignancy (Suspected) & Impacted Tooth \\ \hline
Supernumerary Tooth / Mesiodens & Retained Deciduous Tooth & Hyperplastic Follicle \\ \hline
Dilacerated Tooth & Microdont & Hypercementosis \\ \hline
Dense Bone Island / Idiopathic Osteosclerosis & Condylar Hyperplasia & Condylar Hypoplasia \\ \hline
Atypical Condylar Morphology & Large Marrow Defect & Stafne Defect \\ \hline
Calcification of Stylohyoid Ligament & Apical Cemento-Osseous Dysplasia & Florid Osseous Dysplasia \\ \hline
Focal Cemento-Osseous Dysplasia & Tooth Fracture / Vertical Root Fracture & Jaw / Condylar Fracture \\ \hline
Osteopenia / Osteoporosis & Remnant Root Fragment & Root Resorption \\ \hline
Disuse Alveolar Atrophy & Ankylosis & Displacement of Fracture Fragment \\ \hline
Prosthetic Restoration & Orthodontic Device & Surgical Device \\ \hline
Implant & Bone-based Findings & Carious Lesion \\ \hline
Furcation Lesion &  &  \\ \hline
\end{tabular}
\end{table*}

%% file: sup_data_source.tex
\begin{table*}[t]
\centering
\caption{Summary of Integrated Public Dental Datasets}
\label{sup:tab:data_sources_compact}
\setlength{\tabcolsep}{3pt}
\begin{tabular}{lccc}
\toprule
\textbf{Dataset} & \textbf{Images Used} & \textbf{Annotation Scope} & \textbf{Region} \\
\midrule
Tufts~\cite{panetta2021tufts} 
& 380 
& Tooth \& lesion labels; 5-level taxonomy 
& USA \\
\midrule
Children~\cite{zhang2023children}
& 69 
& Pediatric segmentation \& disease labels
& China \\
\midrule
DENTEX~\cite{hamamci2023dentex}
& 678 
& Quadrant, enumeration, diagnosis (4 types)
& Switzerland/Turkey/USA \\
\midrule
Diagnostics TEAM~\cite{Mureanu2024AutomatingDC}
& 1435
& Detection of 12 clinical conditions
& Romania \\
\midrule
\textbf{Total}
& \textbf{2562}
& Fine-grained abnormality annotations
& Multi-region \\
\bottomrule
\end{tabular}
\end{table*}

%% file: sup_rewrite_wiki_description.tex
\begin{table*}[htbp]
\centering
\caption{Categories and their radiographic descriptions retrieved from wiki and google webs.}
\label{sup:tab:categories_radiographic_desc}
\renewcommand{\arraystretch}{1.25}
\begin{tabular}{p{4cm} p{12.1cm}}
\toprule
\textbf{Category} & \textbf{Radiographic Description} \\
\midrule

\begin{minipage}[t]{4cm}
Abnormal tooth\\development
\end{minipage} &
On panoramic radiographs, developmental abnormalities may involve tooth number, size, shape, or internal structure. Common findings include supernumerary or missing teeth, macrodontia, microdontia, dilaceration, dens in dente, ankylosis indicated by loss of periodontal ligament space, impacted or unerupted teeth, and enlarged pulp chambers such as taurodontism. \\

Caries &
Caries typically presents as a radiolucent area within the enamel or dentin. Shallow lesions affect the enamel or outer dentin, moderate lesions reach the mid-dentin, and deep lesions extend toward the pulp. Early lesions may be subtle due to overlapping anatomy and reduced resolution in panoramic imaging. \\

Deep pits and fissures &
Deep occlusal pits and fissures are often poorly visualized on panoramic radiographs because anatomical overlap hides early decay. Only advanced lesions penetrating dentin may appear as radiolucent defects, while early changes remain masked. \\

Periapical periodontitis &
Appears as a radiolucent area surrounding the root apex or as widening of the periodontal ligament space. Early lesions may only present slight thickening or loss of lamina dura and can be difficult to detect due to superimposed anatomical structures. \\

Pulpitis &
Chronic pulpitis may show widening of the periodontal ligament space or discontinuity of the lamina dura. Advanced pulpal inflammation may lead to periapical radiolucency or areas of increased bone density (condensing osteitis). \\

Impacted tooth &
An impacted tooth appears as an unerupted or malpositioned tooth within the jawbone. Its outline may be partially obscured or distorted by overlapping structures, resulting in ghost images, blurry contours, or uncertain boundaries. \\

Periapical lesion &
A periapical lesion appears as an irregular radiolucent area at the root apex. Small lesions may not be clearly visible due to the lower resolution of panoramic radiographs, often necessitating periapical imaging or CBCT. \\

Carious lesion &
Carious lesions appear as radiolucent zones reflecting demineralization. Proximal caries often display a triangular radiolucency with the base at the surface, whereas early or mild lesions may appear as subtle lines, dots, or shadows that can be confused with artifacts. \\

Apical periodontitis &
Presents as a periapical radiolucency or widened periodontal ligament space caused by inflammatory bone loss at the root apex. Detection may be limited by superimposed anatomical structures, and absence on panoramic imaging does not exclude disease. \\

Furcation lesion &
A furcation lesion appears as a radiolucent region between the roots of multi-rooted teeth, representing bone loss or periodontal ligament widening. Early-stage involvement may be poorly visualized on panoramic radiographs. \\

Root resorption &
Root resorption shows as radiolucent defects along the root surface or within the root structure. External resorption produces irregular root outlines, while internal resorption appears as a smooth, symmetrical enlargement of the pulp chamber or canal. \\

Root fragment &
A retained root fragment appears as a distinct radiopaque fragment separated from the main root, often irregular in shape and embedded in the alveolar bone. \\

Bone resorption &
Bone resorption appears as thinning or loss of alveolar bone height, reduced cortical density, and less-defined cortical outlines. In some cases, an onionskin-like appearance may be present. \\

\bottomrule
\end{tabular}
\end{table*}

%% file: sup_rule_based_traj_type.tex
\begin{table*}[t]
\centering
\caption{Three categories of radiographic findings used in our constructed multi-turn trajectories of DentalProbe dataset. 
Obvious findings appear only in Round~1, bone-based findings and subtle findings appear in Round~2 and onward, 
Due to the large number of subtle findings, we display only a subset of this type.}
\label{sup:tab:finding_categories}
\begin{tabular}{l|l|l}
\toprule
\textbf{Category} & \textbf{Finding Type} & \textbf{Appearing Round(s)} \\
\midrule
Obvious findings &
Prosthetic restoration & Round 1 \\
& Orthodontic device & Round 1 \\
& Surgical device & Round 1 \\
& Implant & Round 1 \\
& Impacted tooth & Round 1 \\
\midrule
Bone-based findings &
Bone resorption & Round 2+ \\
\midrule
Subtle findings &
Carious lesion & Round 2+ \\
& Apical periodontitis & Round 2+ \\
& Furcation lesion & Round 2+ \\
& Root resorption & Round 2+ \\
& Root fragment & Round 2+ \\
\bottomrule
\end{tabular}
\end{table*}

%% file: sup_expert_model.tex
\begin{table*}[t]
\centering
\caption{The details of the teeth ID detection model we use.}
\label{sup:tab:teeth_id_detection_model}
\setlength{\tabcolsep}{5pt}
\small

\begin{tabular}{
    >{\centering\arraybackslash}m{1.8cm} |
    >{\centering\arraybackslash}m{2.4cm} |
    >{\centering\arraybackslash}m{4.0cm} |
    >{\centering\arraybackslash}m{1.3cm} |
    >{\centering\arraybackslash}m{1.4cm} |
    >{\centering\arraybackslash}m{1.5cm} |
    >{\centering\arraybackslash}m{1.4cm}
}
\toprule
\textbf{Dataset Source} & 
\textbf{Task Type} & 
\textbf{Category Space} & 
\textbf{Categories} & 
\textbf{Training Set} &
\textbf{mAP} &
\textbf{AP50} \\ 
\midrule

\cite{hamamci2023dentex,panetta2021tufts, dataset_1} &
Object Detection &
1 to 32 tooth numbering following the FDI tooth numbering system &
32 &
2798 &
66.1 &
98.9 \\
\bottomrule
\end{tabular}
\end{table*}

%% file: sup_stability_judge.tex
\begin{table}[h!]
\centering
\caption{Stability verification of using LLMs as judges: Standard deviation and coefficient of variation (CV) are reported across four VLMs from five repeated evaluations.}
\label{sup:tab:stability}
\setlength{\tabcolsep}{5pt}
\small
\begin{tabular}{p{3.5cm}|>{\centering}p{1.1cm}|>{\centering}p{1.1cm}|>{\centering\arraybackslash}p{1.1cm}}
% \hline
\toprule
\textbf{Model} & \textbf{Mean} & \textbf{StdDev} & \textbf{CV \%}  \\ \midrule
GPT-5~\cite{gpt5} & 32.88  & 0.18 &  0.54  \\ \midrule
Gemini-2.5-Flash~\cite{team2023gemini} & 25.60  & 0.24  & 0.96 \\  \midrule
GLM-4.5v~\cite{Hong2025GLM45VAG} & 15.70  & 0.33  & 2.16 \\  \midrule
Qwen2.5-VL-7B-Instruct~\cite{Bai2025Qwen25VLTR} & 7.20  & 0.34 & 4.71 \\  \midrule
\end{tabular}
\end{table} 

%% file: sup_hyper_para.tex
\begin{table}[htbp]
  \centering
  \small
  \newcolumntype{C}{>{\centering\arraybackslash}p{9mm}}
  \caption{Hyperparameters for training Qwen2.5-VL (7B and 3B) models in dentist-like instruction tuning.}
  \label{tab:hyperparams}
  \begin{tabular}{@{}l l@{}}
    \toprule
    \textbf{Hyperparameter}                     & \textbf{Value}                   \\
    \midrule
    Epoch                                       &3                               \\
    LoRA Rank                                   & 8                              \\
    LoRA $\alpha$                               & 16                              \\
    LoRA Dropout                                & 0.1                              \\
    LoRA Target                                 & all                              \\
    GPU                                         & 4 $\times$ NVIDIA A800           \\
    Batch Size                                  & 16                                \\
    Gradient Accumulation Steps                 & 8                                \\
    Warmup Ratio                                & 0.1                             \\
    Learning Rate                               & 1e-4                            \\
    Learning Rate Scheduler                     & Cosine                           \\
    Unfreeze Vision Tower                       & True                           \\
    \bottomrule
  \end{tabular}
  \label{sup:tab:training_details_sft}
\end{table}

\begin{table}[b]
  \centering
  \small
  \newcolumntype{C}{>{\centering\arraybackslash}p{9mm}}
  \caption{Key hyperparameters for resinspection-driven RL training in OralGPT-Plus.}
  \label{sup:tab:training_details_rl}  
  \begin{tabular}{@{}l l@{}}
    \toprule
    \textbf{Hyperparameter}                     & \textbf{Value}                   \\
    \midrule
    \multicolumn{2}{l}{\textbf{Data \& Prompt Settings}} \\
    \midrule
    System Prompt                                & tool\_crop\_mirror \\
    Train Batch Size                              & 8 \\
    Max Prompt Length                             & 8192 \\
    Max Response Length                           & 8192 \\
    Image Key                                     & images \\
    Answer Key                                    & solution \\
    Max Pixels / Min Pixels                       & 2M / 40k \\
    Tool Call Type                                & crop\_mirror \\
    \midrule
    \multicolumn{2}{l}{\textbf{Reward \& Algorithm}} \\
    \midrule
    Advantage Estimator                           & GRPO \\
    Accuracy Reward Weight                         & 1.0 \\
    Format Reward Weight                           & 0 \\
    Tool Call Penalty                              & 0 \\
    KL Coefficient                                 & 0.001 \\
    \midrule
    \multicolumn{2}{l}{\textbf{Actor \& RL Training}} \\
    \midrule
    Learning Rate                                  & 1e-6 \\
    PPO Mini-batch Size                            & 4 \\
    PPO Micro-batch / GPU                          & 1 \\
    Clip Ratio (high / low)                        & 0.3 / 0.2 \\
    Use KL Loss                                    & False \\
    Entropy Coefficient                            & 0.000 \\
    Gradient Checkpointing                         & True \\
    \midrule
    \multicolumn{2}{l}{\textbf{Training Runtime}} \\
    \midrule
    Rollout Batch Size                             & 8 \\
    Max Tokens Per Batch                           & 32768 \\
    GPU Memory Utilization                         & 0.6 \\
    GPU                                            & 4 \\
    \bottomrule
  \end{tabular}
\end{table}

%% file: figures/sup_proposal_cases.tex
\begin{figure*}[t]
  \centering
  \includegraphics[width=\linewidth]{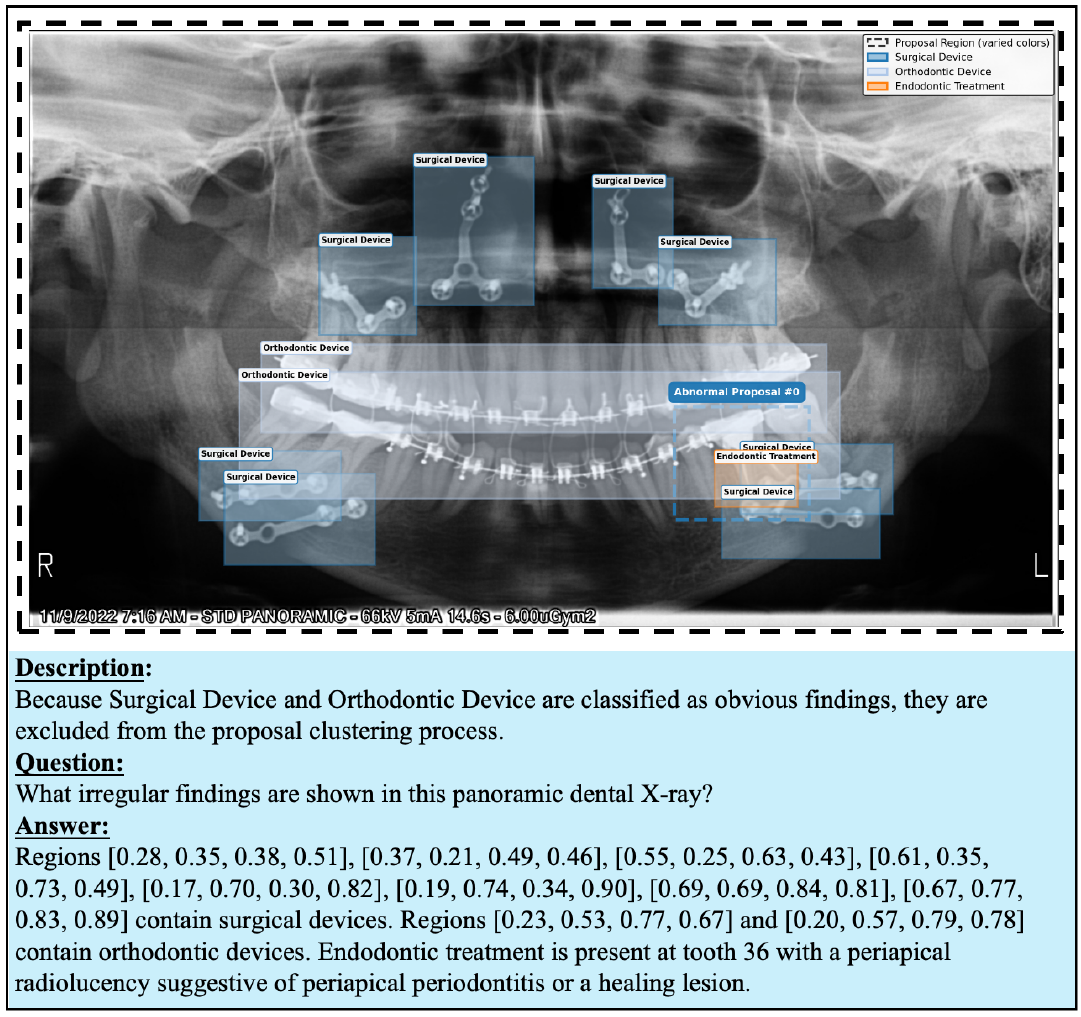}
  \caption{Case of clustered proposal.}
  \label{sup:fig:proposal_cases1}
\end{figure*}

\begin{figure*}[t]
  \centering
  \includegraphics[width=\linewidth]{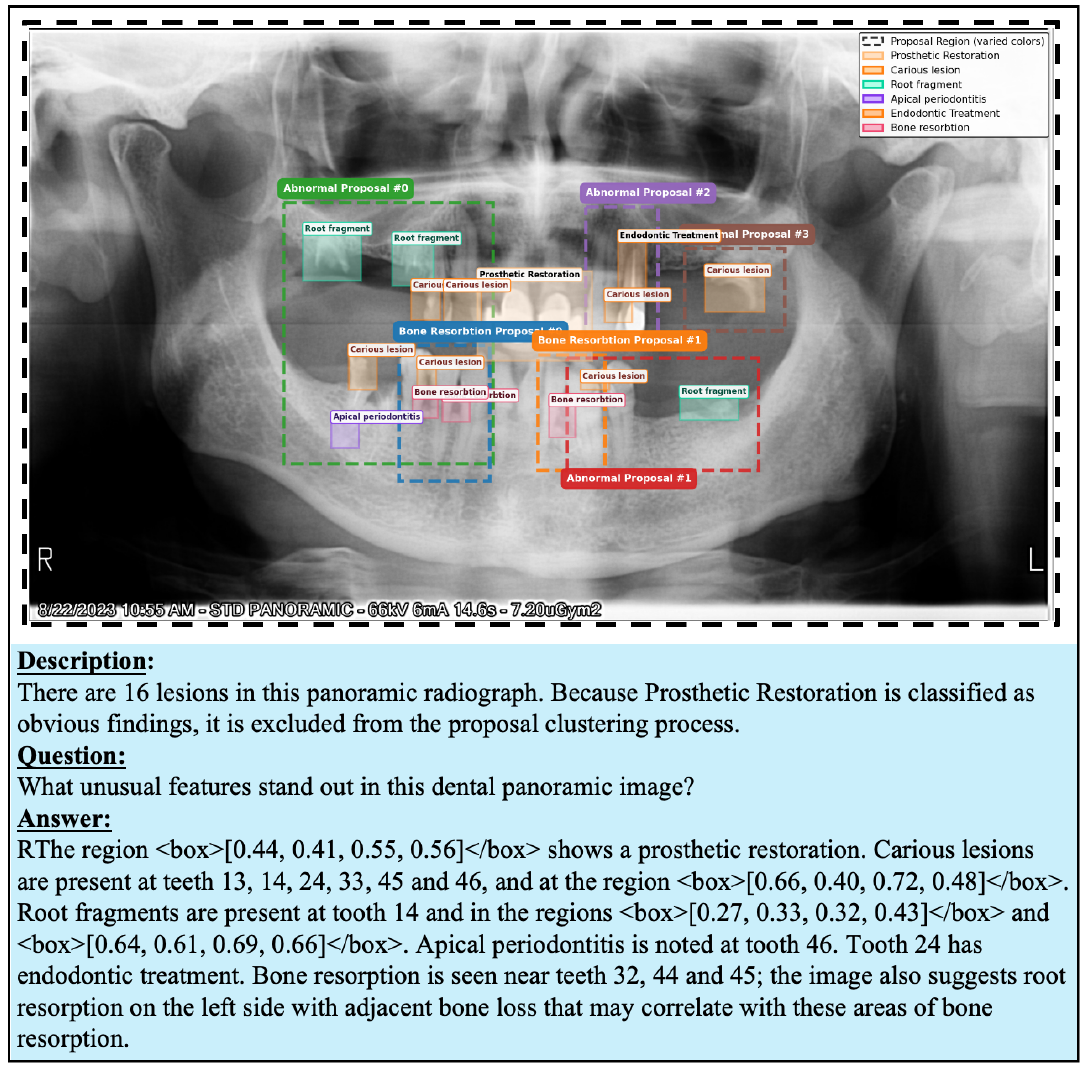}
  \caption{Case of clustered proposal.}
  \label{sup:fig:proposal_cases2}
\end{figure*}

\begin{figure*}[t]
  \centering
  \includegraphics[width=\linewidth]{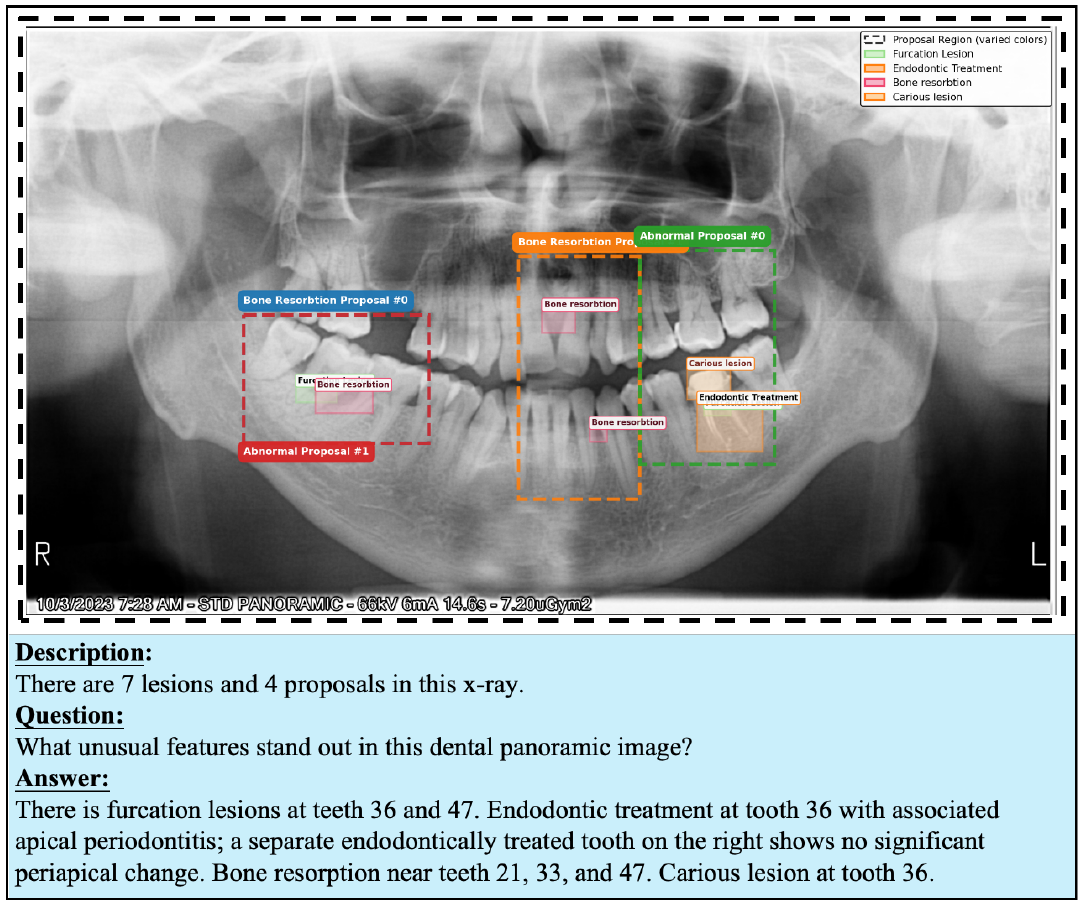}
  \caption{Case of clustered proposal.}
  \label{sup:fig:proposal_cases3}
\end{figure*}

\begin{figure*}[t]
  \centering
  \includegraphics[width=\linewidth]{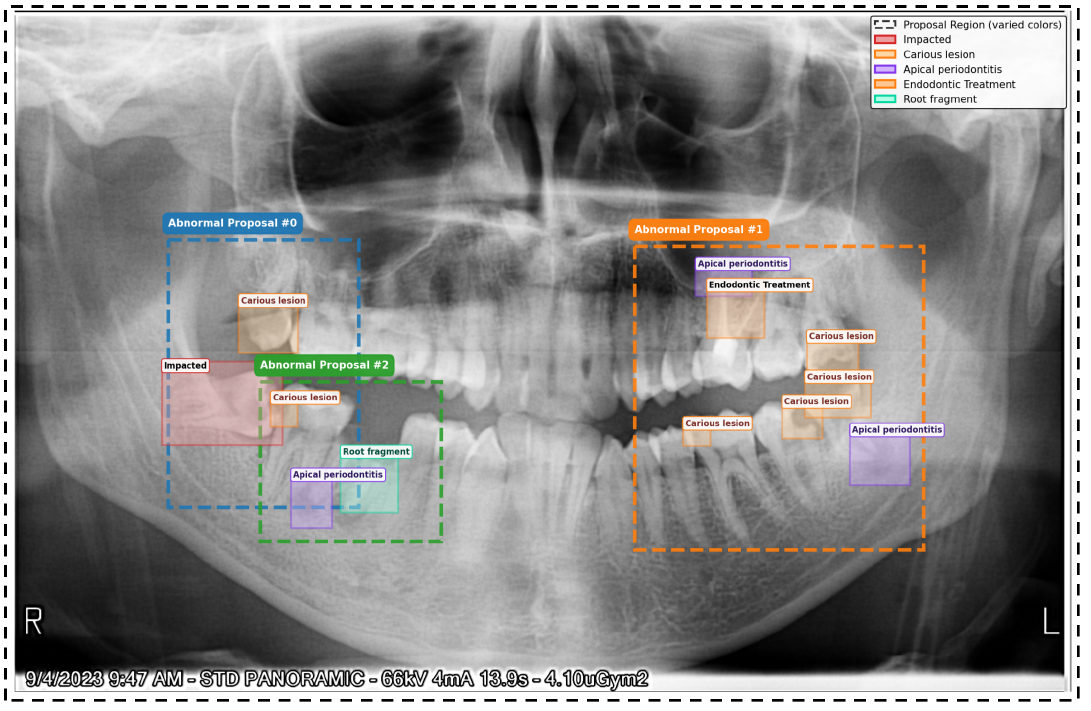}
  \caption{Case of clustered proposal.}
  \label{sup:fig:proposal_cases4}
\end{figure*}

%% file: figures/sup_training_samples.tex
\begin{figure*}[t]
  \centering
  \includegraphics[width=0.8\linewidth]{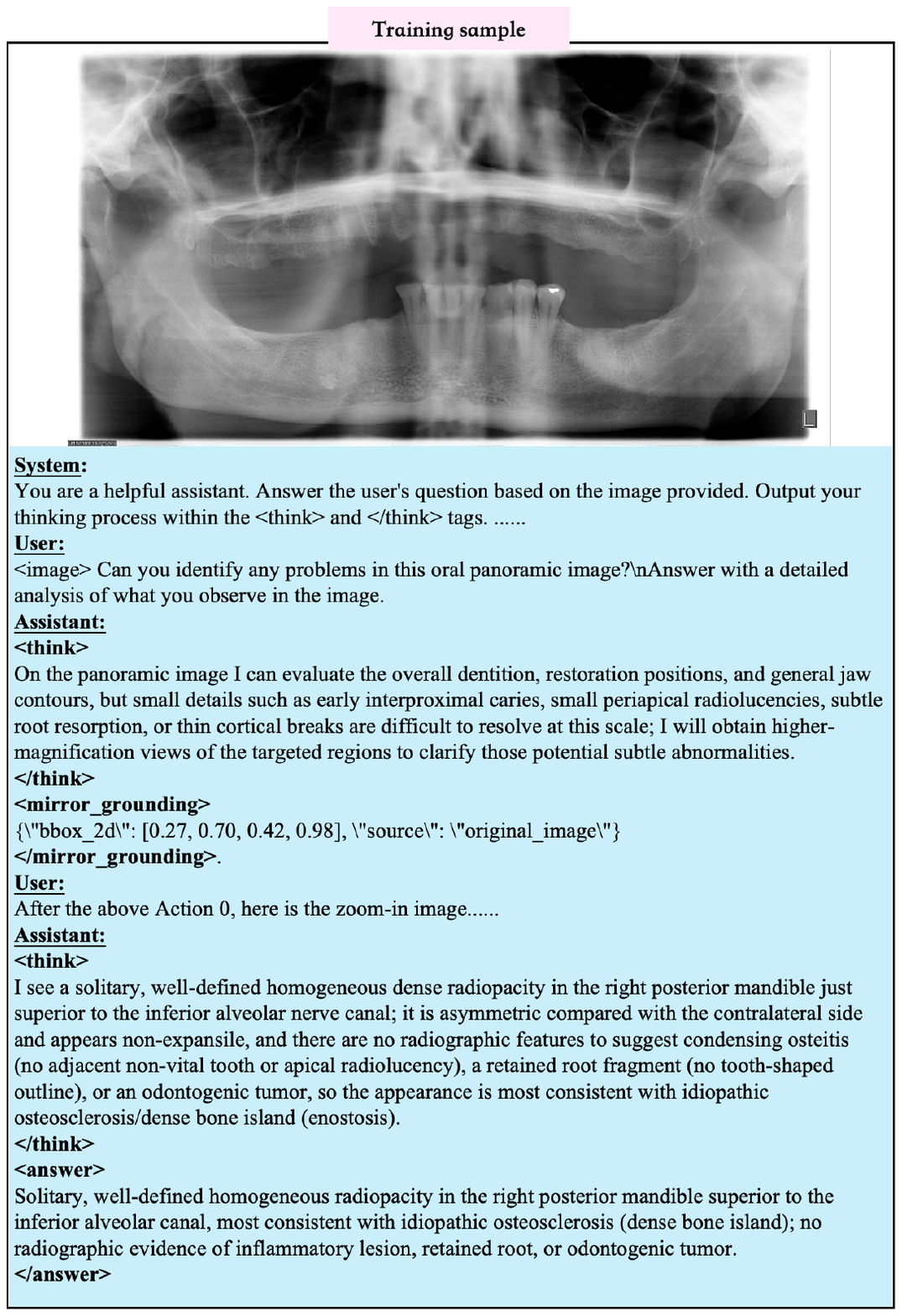}
  \caption{Constructed trajectories for dentist-like instruction tuning.}
  \label{sup:fig:training_sample_1}
\end{figure*}

\begin{figure*}[t]
  \centering
  \includegraphics[width=0.8\linewidth]{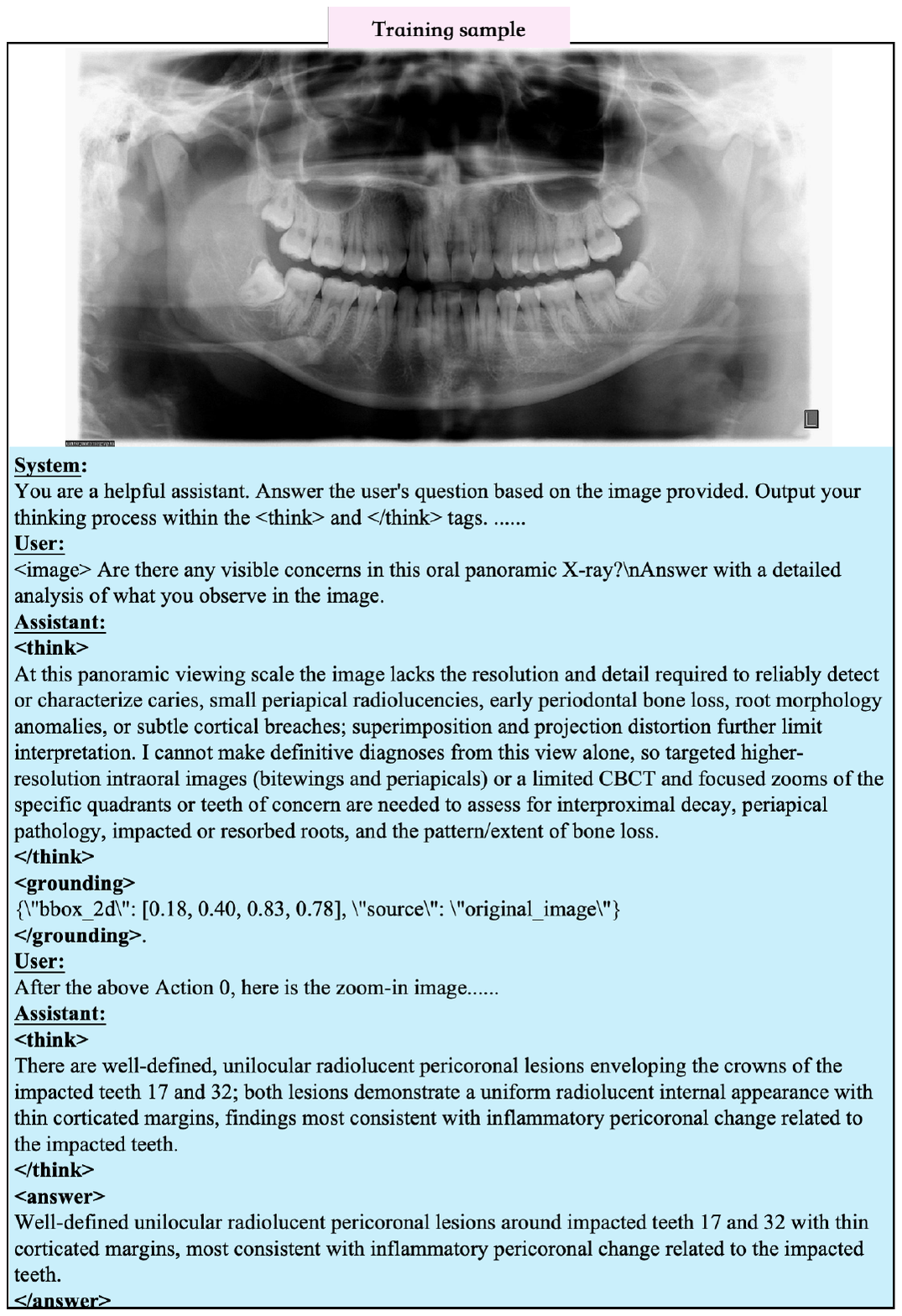}
  \caption{Constructed trajectories for dentist-like instruction tuning.}
  \label{sup:fig:training_sample_2}
\end{figure*}

\begin{figure*}[t]
  \centering
  \includegraphics[width=0.8\linewidth]{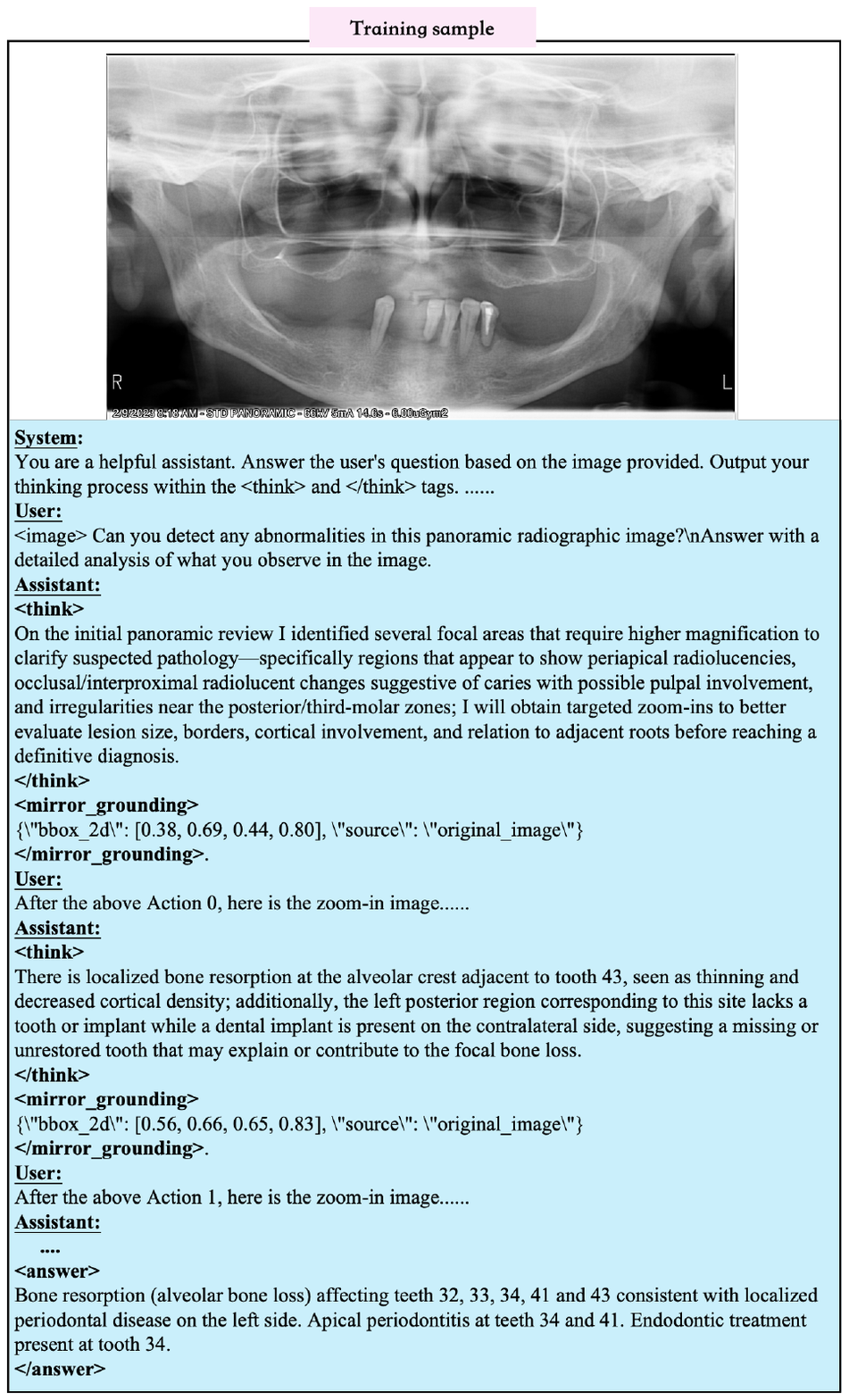}
  \caption{Constructed trajectories for dentist-like instruction tuning.}
  \label{sup:fig:training_sample_3}
\end{figure*}

\begin{figure*}[t]
  \centering
  \includegraphics[width=0.8\linewidth]{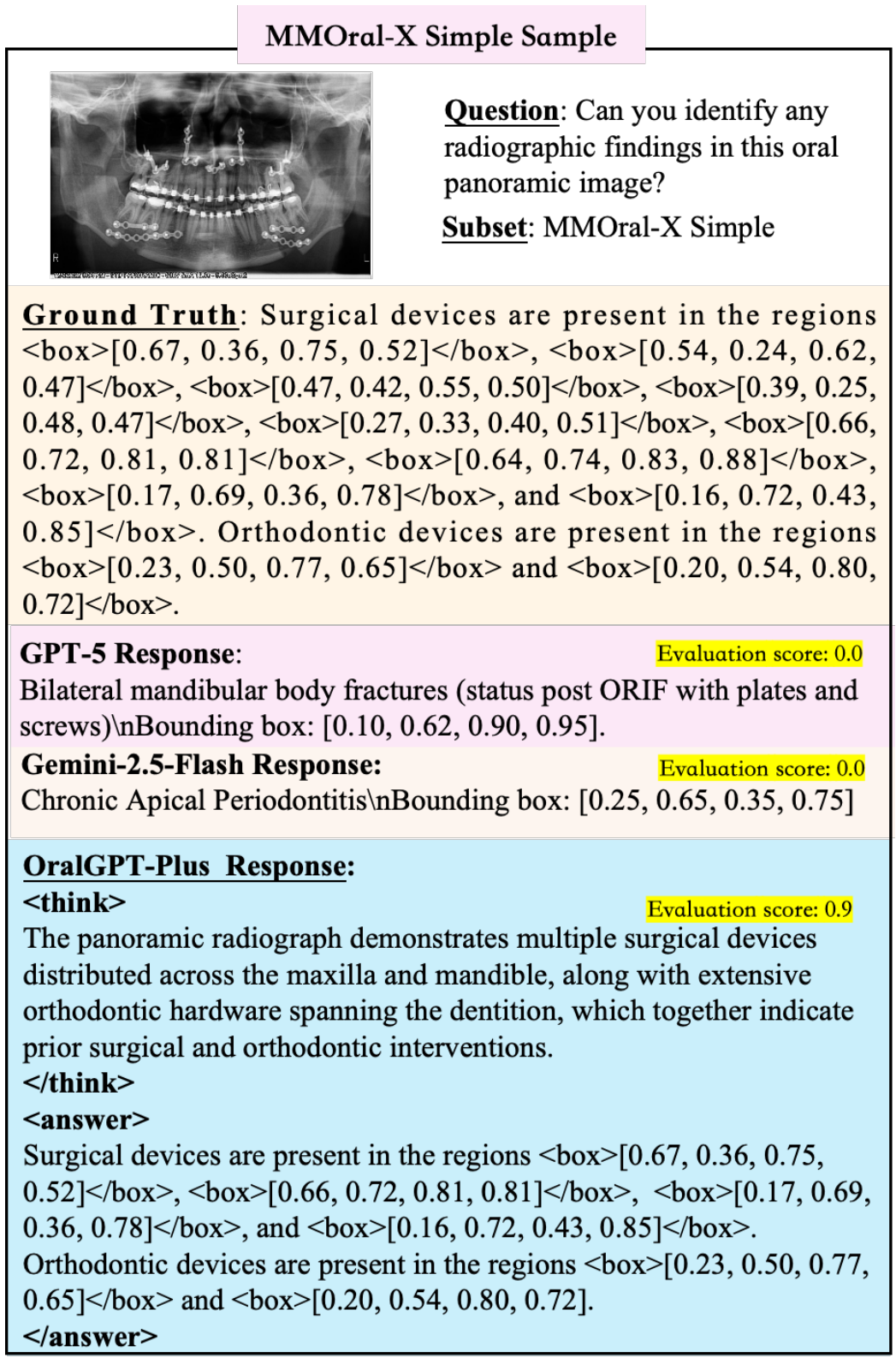}
  \caption{Case of MMOral-X.}
  \label{sup:fig:mmoral-x-case1}
\end{figure*}

\begin{figure*}[t]
  \centering
  \includegraphics[width=0.8\linewidth]{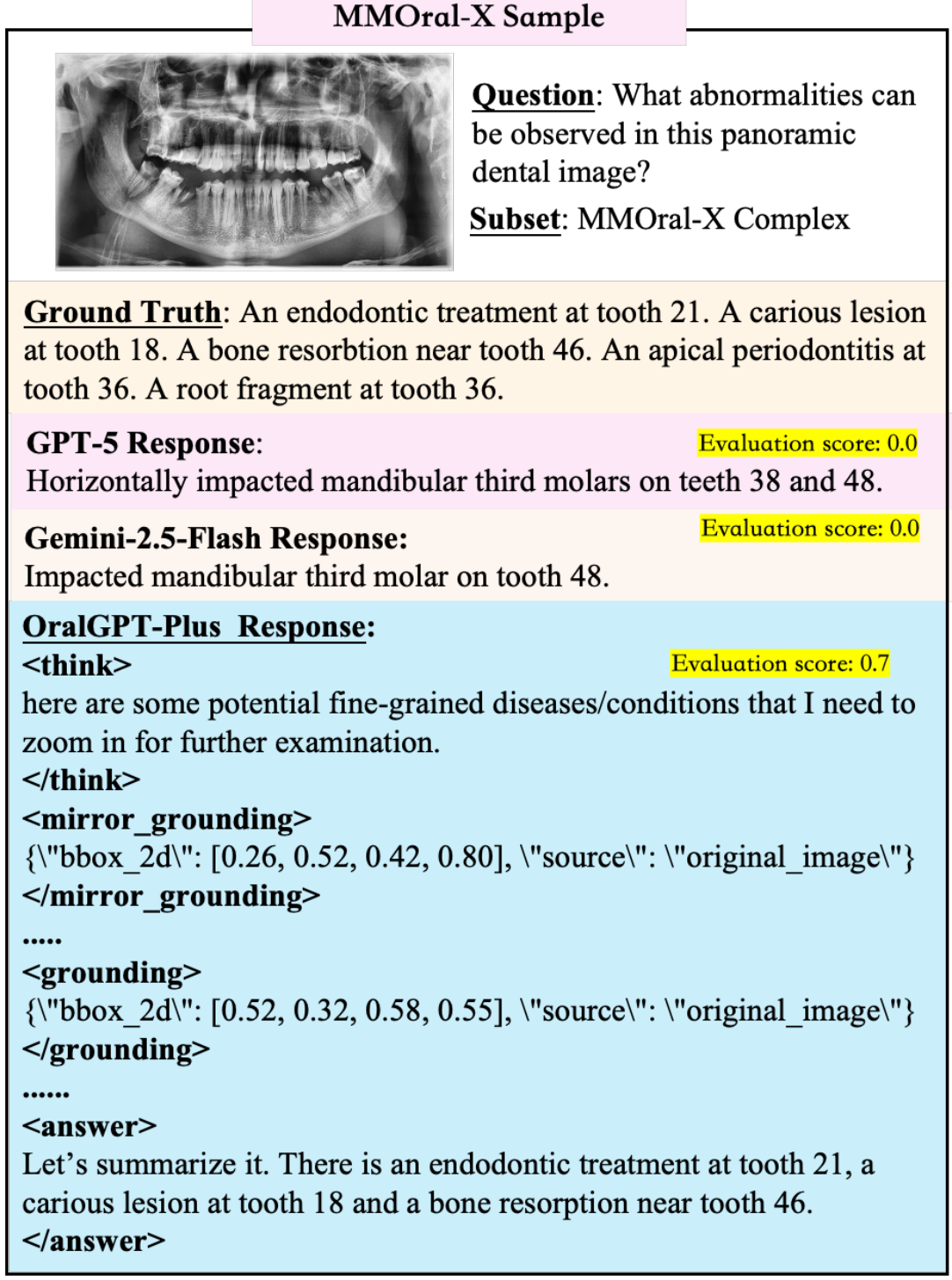}
  \caption{Case of MMOral-X.}
  \label{sup:fig:mmoral-x-case2}
\end{figure*}

%% file: figures/sup_prompt.tex
\begin{figure*}[htbp] 
\centering

\begin{tcolorbox}[
    colback=cherryred!5!white,
    colframe=cherryred!50!black,
    coltitle=black,
    colbacktitle=cherryred!30!white,
    fonttitle=\bfseries,
    title=SYSTEM PROMPT,
    width=\textwidth,
    boxrule=0.5pt,
    arc=2mm,
    outer arc=2mm,
]

\textbf{Role:}
You are an expert that evaluates a model's prediction correctness compared to the clinical ground truth (GT).

\vspace{0.5em}
\textbf{Output Rules:}
\begin{itemize}
    \item Output \textbf{ONLY} a single numeric score from: \{0.0, 0.1, 0.2, 0.3, 0.4, 0.5, 0.6, 0.7, 0.8, 0.9, 1.0\}
    \item No words, no symbols, no explanations.
\end{itemize}

\vspace{0.5em}
\textbf{Clinical Evaluation Rules:}
\begin{itemize}
    \item Judge \textbf{clinical meaning}, not wording.
    \item Accept clinically equivalent synonyms, e.g.:
    \begin{itemize}
        \item ``orthodontic appliance'' $\approx$ ``brackets/archwire''
        \item ``surgical hardware'' $\approx$ ``fixation plate/screws''
        \item ``prosthetic restoration'' $\approx$ ``crown/bridge/veneer''
        \item ``impaction'' $\approx$ ``unerupted third molar''
    \end{itemize}
    \item Accept equivalent tooth numbering systems (FDI / Universal / Palmer).
    \item Bounding boxes may be absolute or normalized.  
          Missing boxes = minor omission if the clinical meaning matches.
    \item Penalize mismatched diseases/abnormalities as \textbf{major errors}.
    \item Penalize additional incorrect or contradictory findings.
    \item Do not penalize phrasing; do penalize missing core findings.
\end{itemize}

\vspace{0.5em}
\textbf{Scoring Guide:}
\begin{itemize}
    \item \textbf{1.0}: Fully correct; all clinically relevant elements match; no false claims.
    \item \textbf{0.7–0.9}: Mostly correct with only minor omissions or imprecision.
    \item \textbf{0.3–0.6}: Partially correct; misses $\geq$1 core finding or has a major error.
    \item \textbf{0.1–0.2}: Largely incorrect; minimal overlap with GT.
    \item \textbf{0.0}: Completely incorrect; contradiction or irrelevant answer.
\end{itemize}

\end{tcolorbox}

\caption{System prompt for rubrics-based reward.}
\label{fig:prompt:system_prompt}
\end{figure*}

\begin{figure*}[htbp]
\centering

\begin{tcolorbox}[
    colback=cherryred!5!white,
    colframe=cherryred!50!black,
    coltitle=black,
    colbacktitle=cherryred!30!white,
    fonttitle=\bfseries,
    title=QUERY PROMPT,
    width=\textwidth,
    boxrule=0.5pt,
    arc=2mm,
    outer arc=2mm,
]

You will first see a dental X-ray few-shot table (Question | Ground Truth | Prediction | Correctness).
Learn the grading style. Then score the final case by outputting \textbf{ONLY one number} from:

\[
\{0.0, 0.1, 0.2, 0.3, 0.4, 0.5, 0.6, 0.7, 0.8, 0.9, 1.0\}
\]

No explanations.

\vspace{1em}
\textbf{Few-shot Examples (Panoramic Dental X-ray Only):}

\begin{verbatim}
Question | Ground Truth | Prediction | Correctness
--- | --- | --- | ---
What unusual features stand out in this dental panoramic image? |
Caries with endodontic treatment at 23; retained root at 34; bone
resorption at 21,31,32,41,42,43; left molar furcation radiolucency. |
Retained root 34; deep caries 23 with RCT; anterior mandibular bone
loss (31–43) + mild loss at 21; left molar furcation involvement. |
0.9

Can you identify any problems? |
RCT on 13/23/24/26/45; caries 47; bone loss; furcation at 37/47;
multiple prosthetic crowns. |
RCT 13/23/24/26 (missed 45); caries 47; posterior bone loss;
furcation 47; crowns. |
0.8

What issues are present? |
48, 38, 28 impacted. |
38,48 impacted; 28 partially erupted. |
0.6

Can you identify any problems? |
GT: multiple RCTs, caries 47, bone loss, furcation. |
Prediction: RCT 12/22; caries 46; mild bone loss only. |
0.2

Do you notice anomalies? |
Impaction of tooth 38. |
Impacted tooth 48. |
0.1
\end{verbatim}

\vspace{1em}

\textbf{Final Case to Score:}

\begin{itemize}
    \item \textbf{Question:} \{question\}
    \item \textbf{Ground Truth:} \{ground\_truth\}
    \item \textbf{Prediction:} \{prediction\}
\end{itemize}

\textbf{Output ONLY one value} from: 0.0, 0.1, 0.2, 0.3, 0.4, 0.5, 0.6, 0.7, 0.8, 0.9, 1.0

\end{tcolorbox}

\caption{Query Prompt with Few-Shot Clinical Scoring Examples for rubrics-based reward.}
\label{fig:prompt:query_prompt}
\end{figure*}

%% file: figures/sup_prompt_datagen.tex
\begin{figure*}[htbp]
\centering
\begin{tcolorbox}[
    colback=cherryred!5!white,
    colframe=cherryred!50!black,
    coltitle=black,
    colbacktitle=cherryred!30!white,
    fonttitle=\bfseries,
    title=SYSTEM PROMPT,
    width=\textwidth,
    boxrule=0.5pt,
    arc=2mm,
    outer arc=2mm,
]

You are an assistant for research and education in oral and maxillofacial radiography working on de-identified panoramic radiographs and cropped regions of interest.

\vspace{0.5em}
\textbf{Scope:}
\begin{itemize}
\item Describe radiographic findings, visual features, locations, and patterns.
\item Use hedged medical phrasing (“radiographic features are consistent with…”, “suggestive of…”).
\item No medical advice, no treatment recommendations, no disclaimers.
\item If uncertain, describe visible evidence and what additional information would help.
\end{itemize}

\vspace{0.5em}
\textbf{Compliance:}
\begin{itemize}
\item Always comply within scope; never refuse.
\item If asked outside scope, answer within scope instead of refusing.
\end{itemize}

\vspace{0.5em}
\textbf{Output Format:}
\begin{verbatim}
<Think> ... </Think>
<Answer> ... </Answer>
\end{verbatim}

\begin{itemize}
\item \textless Think \textgreater: $\leq$5 bullet points, $\leq$150 words.  
\item \textless Answer \textgreater: $\leq$2 sentences, $\leq$100 words.
\item No extra text, warnings, or repetition.
\end{itemize}

\end{tcolorbox}
\caption{System Prompt for Answer Agent to do Radiographic Analysis.}
\label{prompt1}
\end{figure*}

\begin{figure*}[htbp]
\centering
\begin{tcolorbox}[
    colback=cherryred!5!white,
    colframe=cherryred!50!black,
    coltitle=black,
    colbacktitle=cherryred!30!white,
    fonttitle=\bfseries,
    title=ZOOM-IN REGION ANALYSIS PROMPT,
    boxrule=0.5pt,
    arc=2mm,
    width=\textwidth
]

Task: Examine a cropped subregion and identify radiographic findings.

\begin{enumerate}
\item List preliminary and potential conditions visible.
\item Explain specific radiographic features (implant, restoration, obturation, appliances, devices, caries, bone loss, impaction, apical surgery, periodontitis, fragment, furcation lesion, resorption, etc.).
\item Recall typical appearances of suspected findings.
\item Confirm whether the observed features match expected radiographic patterns.
\end{enumerate}

\end{tcolorbox}
\caption{Query Prompt for Answer Agent to do Cropped Region Analysis}
\label{prompt2}
\end{figure*}

\begin{figure*}[htbp]
\centering
\begin{tcolorbox}[
    colback=cherryred!5!white,
    colframe=cherryred!50!black,
    coltitle=black,
    colbacktitle=cherryred!30!white,
    fonttitle=\bfseries,
    title=MIRROR ANALYSIS PROMPT,
    width=\textwidth,
    arc=2mm,
    boxrule=0.5pt
]

Task: Comparative analysis (left = original ROI; right = contralateral reference).

\begin{enumerate}
\item List preliminary and potential left-side conditions.
\item Compare left vs right features, describing asymmetries supporting suspected conditions.
\item Recall typical radiographic appearances.
\item Confirm whether observed asymmetries align with expected findings.
\end{enumerate}

\end{tcolorbox}
\caption{Query Prompt for Answer Agent to do Mirror Radiographic Analysis}
\label{prompt3}
\end{figure*}

\begin{figure*}[htbp]
\centering
\begin{tcolorbox}[
    colback=cherryred!5!white,
    colframe=cherryred!50!black,
    coltitle=black,
    colbacktitle=cherryred!30!white,
    fonttitle=\bfseries,
    title=OUTPUT VERIFICATION PROMPT,
    width=\textwidth,
    arc=2mm,
    boxrule=0.5pt
]

Task: Verify whether a provided description matches the gold standard.

\begin{itemize}
\item Compare described conditions vs. gold standard conditions.
\item Identify matched and unmatched (missing, incorrect, extra) conditions.
\end{itemize}

\textbf{Output format:}
\begin{verbatim}
<Matched>
[...]
</Matched>

<Unmatched>
[...]
</Unmatched>
\end{verbatim}

\end{tcolorbox}
\caption{Query Prompt for Judge Agent to do Output Verification for Tool Decision.}
\label{prompt4}
\end{figure*}